\RequirePackage{fix-cm}
\documentclass[twocolumn]{svjour3}          
\smartqed  
\usepackage{graphicx}
\usepackage{enumerate}

\usepackage{epsfig}
\usepackage{graphicx}
\usepackage{amsmath}
\usepackage{amssymb}
\usepackage{multirow}
\usepackage{array}
\usepackage{algorithm}
\usepackage{algorithmic}
\usepackage{color}
\usepackage{xcolor}
\usepackage{marvosym}
\usepackage{caption}

\usepackage[colorlinks,
            linkcolor=blue,
            anchorcolor=blue,
            breaklinks,
            citecolor=blue]{hyperref}
 \usepackage{natbib}
\usepackage{etoolbox}
\makeatletter
\patchcmd{\NAT@citex}
  {\@citea\NAT@hyper@{%
     \NAT@nmfmt{\NAT@nm}%
     \hyper@natlinkbreak{\NAT@aysep\NAT@spacechar}{\@citeb\@extra@b@citeb}%
     \NAT@date}}
  {\@citea\NAT@nmfmt{\NAT@nm}%
   \NAT@aysep\NAT@spacechar\NAT@hyper@{\NAT@date}}{}{}

\patchcmd{\NAT@citex}
  {\@citea\NAT@hyper@{%
     \NAT@nmfmt{\NAT@nm}%
     \hyper@natlinkbreak{\NAT@spacechar\NAT@@open\if*#1*\else#1\NAT@spacechar\fi}%
       {\@citeb\@extra@b@citeb}%
     \NAT@date}}
  {\@citea\NAT@nmfmt{\NAT@nm}%
   \NAT@spacechar\NAT@@open\if*#1*\else#1\NAT@spacechar\fi\NAT@hyper@{\NAT@date}}
  {}{}
\makeatother

\begin{document}

\title{Population-Based Evolutionary Gaming for Unsupervised Person Re-identification
}


\author{Yunpeng Zhai         \and
        Peixi Peng     \and  
        Mengxi Jia     \and 
        Shiyong Li     \and
        Weiqiang Chen  \and
        Xuesong Gao    \and
        Yonghong Tian  
}


\institute{\hspace*{1.5em} Yunpeng Zhai$^1$ \at \hspace*{1.5em} ypzhai@pku.edu.cn
          \and
          \Letter \hspace*{0.5em} Peixi Peng$^{1,3}$ \at \hspace*{1.5em} pxpeng@pku.edu.cn
          \and
          \hspace*{1.5em} Mengxi Jia$^1$ \at \hspace*{1.5em} 
          mxjia@pku.edu.cn
          \and
          \hspace*{1.5em} Shiyong Li$^4$ \at \hspace*{1.5em} 
          lishiyong@huawei.com
          \and
          \hspace*{1.5em} Weiqiang Chen$^5$ \at \hspace*{1.5em} 
          chenweiqiang@iCloud.com
          \and
          \hspace*{1.5em} Xuesong Gao$^5$ \at \hspace*{1.5em} 
          xuesong@outlook.com
          \and
          \Letter \hspace*{0.5em} Yonghong Tian$^{2,1,3}$ \at \hspace*{1.5em} yhtian@pku.edu.cn
          \and
          $^1$\hspace*{1em} National Engineering Research Center of Visual \\ 
          \hspace*{1.4em}  Technology, School of Computer Science, \\
          \hspace*{1.4em} Peking University, China.  \at
          \and
          $^2$\hspace*{1em} School of Electronic and Computer Engineering, \\
          \hspace*{1.4em} Peking University Shenzhen Graduate School. \at
          \and
          $^3$\hspace*{1em} Peng Cheng Laboratory, China. \at
          \and
          $^4$\hspace*{1em} AI Application Research Center, Huawei \\ \hspace*{1.4em} Technologies Co., Ltd, China. \at
          \and
          $^5$\hspace*{1em} State Key Laboratory of Digital Multimedia \\
          \hspace*{1.4em} Technology,  Hisense, Qingdao, China. \at
}

\date{Received: date / Accepted: date}

\maketitle

\begin{abstract}
Unsupervised person re-identification has achieved great success through the self-improvement of individual neural networks. However, limited by the lack of diversity of discriminant information, a single network has difficulty learning sufficient discrimination ability by itself under unsupervised conditions. To address this limit, we develop a population-based evolutionary gaming (PEG) framework in which a population of diverse neural networks are trained concurrently through selection, reproduction, mutation, and population mutual learning iteratively. Specifically, the selection of networks to preserve is modeled as a cooperative game and solved by the best-response dynamics, then the reproduction and mutation are implemented by cloning and fluctuating hyper-parameters of networks to learn more diversity, and population mutual learning improves the discrimination of networks by knowledge distillation from each other within the population. In addition, we propose a cross-reference scatter (CRS) to approximately evaluate re-ID models without labeled samples and adopt it as the criterion of network selection in PEG. CRS measures a model's performance by indirectly estimating the accuracy of its predicted pseudo-labels according to the cohesion and separation of the feature space. Extensive experiments demonstrate that (1) CRS approximately measures the performance of models without labeled samples; (2) and PEG produces new state-of-the-art accuracy for person re-identification, indicating the great potential of population-based network cooperative training for unsupervised learning. 

\end{abstract}

\begin{figure*}[t]
\centering
\includegraphics[width=1\linewidth]{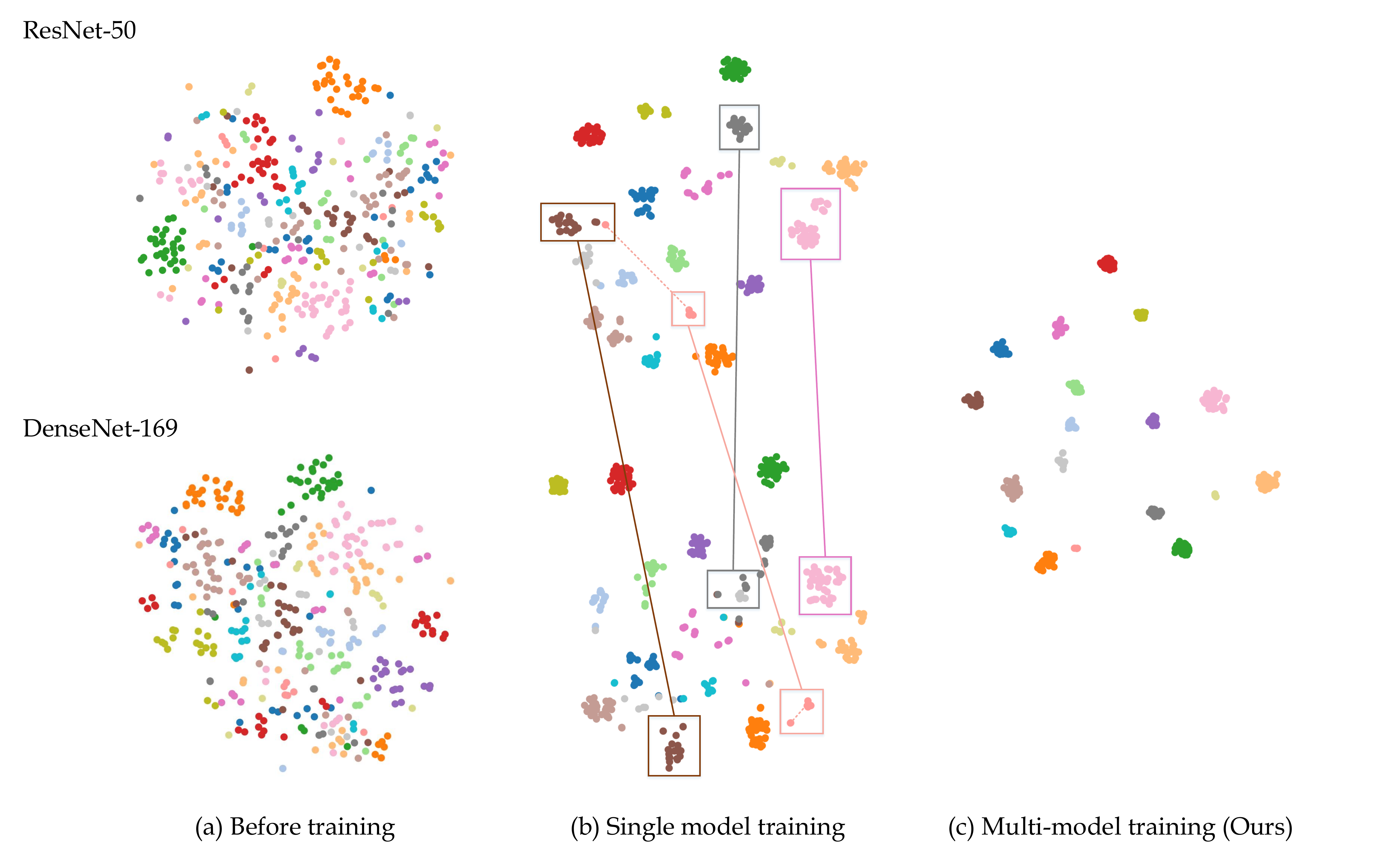}
\caption{Feature distribution of the same samples with different methods where each color denotes a person identity. Single model training(b) uses the self-learning mechanism only to enhance the discrimination ability it already has before training(a) and still suffers from inaccurate pseudo-labels. However, multi-model training(c) explores and exploits the complementary information among different models (marked by corresponding colored boxes) and achieves more discrimination.}
\label{fig:motivation}
\end{figure*}

\section{Introduction}
Person re-identification (re-ID) aims to match persons in an image gallery collected from non-overlapping camera networks, which has attracted increasing interest thanks to its wide applications in security and surveillance. Though supervised re-ID methods ~\citep{yang2020part}~\citep{zheng2016person} have achieved very decent results, they are largely dependent on sufficient data with expensive manual annotation, which also require substantial personal identity information and entail privacy issues. By contrast, unsupervised re-ID not only reduces the cost of labeling but also protects personal privacy without checking images manually. 
Commonly, unsupervised re-ID can be divided into two categories: unsupervised domain adaptation (UDA) ~\citep{zhai2020ad}~\citep{zhong2020learning} and fully unsupervised re-ID (FU)~\citep{chen2021ice}~\citep{lin2019bottom} depending on whether using extra labeled data. 
In this study, we will mainly focus on the fully unsupervised setting which learns directly from unlabeled images and allows for more scalability in real-world deployments.

To address the challenges of unsupervised re-ID, recent efforts concentrate on training individual neural networks by means of a self-improvement strategy ~\citep{DBLP:journals/corr/abs-1807-11334}~\citep{ge2020self}. They attempt to learn better representations based on self-predicted pseudo-labels via clustering algorithms ~\citep{caron2018deep} or graph neural networks ~\citep{ye2017dynamic}. 
{However, a single model can use such a self-learning mechanism only to enhance the discrimination ability it already has and cannot tackle the incorrectly predicted pseudo-labels, which prevents it from maximizing its discrimination. Due to the lack of diversity of single models, incorrect pseudo-labels are likely to remain the same after unsupervised training such as the false positive samples where images of different persons are clustered into the same group or the false negative samples where the images of the same person are clustered into different groups, as shown in Fig. \ref{fig:motivation}. Importantly, since models learn diverse discrimination with different architectures, the incorrect pseudo-labels predicted by a model may be predicted correctly by another model, marked by boxes in Fig. \ref{fig:motivation}(b). 
In this paper, we attempt to address unsupervised re-ID by multiple model training, in which the complementary information of different models can be integrated and utilized effectively to explore the various latent knowledge contained in unlabeled data (the quantitative analysis is shown in Sec. \ref{sec:ab-components}).

}

However, multiple model training still faces two challenging issues: {(1) How to learn diverse discrimination with multiple different models? (2) How to select a set of better models from many diverse models for training?} To tackle these issues, we propose a population-based evolutionary gaming (PEG), which selects and trains discriminative models by exploration and exploitation of their diversity. 
PEG trains a population of models concurrently by iterative selection, reproduction, mutation, and population mutual learning of neural networks, as shown in Fig. \ref{fig:overview}. 
Specifically, selection adapts the whole population to the unlabeled data by selecting and preserving the optimal subset of networks with complementary discrimination ability while abandoning other networks out of the subset. This combinatorial optimization of networks in selection is modeled as a multi-agent cooperative game and solved by the best response dynamics, in which each agent attempts to learn the best response to the other agents’ action and thus leads to Nash equilibrium. Then, reproduction and mutation are performed on the selected population to increase its diversity by making multiple copies of each network and applying a stochastic disturbance to their hyper-parameters. Selection and reproduction jointly maintain the size of the population.
Afterward, population mutual learning is conducted among networks to assemble and further explore the discrimination capacity via knowledge distillation within populations. Each network learns representations from both population-shared pseudo-labels and soft-labels predicted by other individual networks. Utilizing periodically performing selection, reproduction and mutation, population mutual learning, the evolutionary gaming process enables favorable traits and knowledge of neural networks to be transmitted through successive generations.

In the evolution gaming, a core issue is to define the utility function of the game, that is, the criterion of network selection in the evolution. However, the evaluation of CNN models without labeled datasets has not been well studied. Here, we propose cross-reference scatter (CRS), which can approximately evaluate the quality of networks using unlabeled samples. Generally, the pseudo-labels predicted by better networks are more accurate; however, their accuracy cannot be directly evaluated when the ground truth is unavailable. Moreover, models trained by more accurate pseudo-labels tend to achieve larger intra-cluster cohesion and inter-cluster separation in the feature space because incorrect labels will enforce models to separate samples of the same class or aggregate samples of different classes. Motivated by this phenomenon, we indirectly evaluate a network according to the feature cohesion and separation of a reference model that is trained by pseudo-labels of the evaluated network. Hence, the CRS is defined by the ratio of the inter-cluster and intra-cluster variance of features to measure both separation and cohesion. We demonstrate that the CRS approximately reflects the discrimination capacity of models without ground truth data and thus promotes the evolution gaming to learn better representations.

A preliminary version of this work has been partially published ~\citep{zhai2020multiple}, which has demonstrated the effectiveness of mutual learning among multiple networks in unsupervised conditions. Based on that version, this manuscript has made great improvements, including: 1) We propose a novel population-based evolutionary gaming (PEG) framework (Sec. \ref{sec:peg}). The previous algorithm works passively only on given networks, and cannot adaptively select the most suitable models from the model base. Based on the mutual learning, PEG additionally contains an iterative selection of networks via a multi-agent cooperative game preventing the weak networks to distract the overall discrimination capability (Sec. \ref{sec:sel}). 2) We propose a new cross-reference scatter (CRS) to approximately measure re-ID models without labeled data. To evaluate the model discrimination, the previous version introduced inter-/intra-cluster scatter to roughly modulate the weights of models during mutual learning. However, it cannot be considered as the utility function of the cooperative game in PEG due to the lack of capability to accurately evaluate models. This paper improves inter-/intra-cluster scatter to cross-reference scatter by adding a cross-reference evaluation (CR) scheme (Sec. \ref{sec:crs}). 3) More qualitative and quantitative experiments are conducted to evaluate the effectiveness of the method, including but not limited to the validation and analysis of CRS, the cooperative game, and PEG. 

In summary, our contribution is as follow:
\begin{itemize}
\item It proposes a novel population-based evolutionary gaming framework for unsupervised person re-ID which trains a diverse population of neural networks by iterative selection, reproduction, mutation and mutual learning.

\item It introduces a multi-agent cooperative game for the selection of networks in the PEG, which aims to find and preserve an optimal subset of the population on unlabeled data. 

\item It investigates the evaluation of re-ID models using unlabeled data and proposes a cross-reference scatter which approximately measures a model's discrimination capability by indirectly estimating its predicted pseudo-labels according to the cohesion and separation of feature space.

\item Experiments show that PEG outperforms state-of-the-art methods on large-scale datasets, indicating the great potential of population-based multiple model training.

\end{itemize}

\section{Related Works}
\subsection{Unsupervised Person Re-ID}
Unsupervised person re-ID can be categorized into Unsupervised Domain Adaptation (UDA) and Fully Unsupervised Re-ID (FU).
UDA methods try to train a re-ID model by unlabeled target data together with labeled source data, while FU methods attempt to train models with only unlabeled data after pre-training. Despite the different data conditions, most UDA and FU methods adopt similar learning strategies which can be summarized into two categories. A line of works are mainly based on alignment to reduce distribution shift between cameras or domains in pixel level, such as SPGAN~\citep{Deng_2018_CVPR}, CamStyle~\citep{DBLP:journals/tip/ZhongZZLY19}, HHL~\citep{DBLP:conf/eccv/ZhongZLY18}, ECN~\citep{Zhong_2019_CVPR}, ATNet~\citep{Liu_2019_CVPR}, PDA-Net~\citep{li2019cross}, DG-Net++~\citep{zou2020joint} and GCL~\citep{chen2021joint}, or feature level, such as TJ-AIDL~\citep{Wang_2018_CVPR}, DAAM~\citep{huang2019domain}, UCDA-CCE~\citep{qi2019novel} IICS~\citep{xuan2021intra} and CAP~\citep{wang2020camera}. This line of methods sufficiently utilize the reliable information of camera or domain styles but ignore the latent relationship among unlabeled samples, which hinders them from better performance. 
Another line of works are based on pseudo label discovery, which rely on the iteration of pseudo-label mining and model fine-tuning ~\citep{DBLP:journals/tomccap/FanZYY18}~\citep{DBLP:journals/corr/abs-1807-11334}~\citep{zhang2019self}~\citep{jin2020global}~\citep{zhao2020unsupervised}~\citep{zheng2021online}, such as BUC~\citep{lin2019bottom}, SSG~\citep{fu2019self}, ~\citep{Zhai_2020_CVPR}, HCT~\citep{zeng2020hierarchical} and SpCL~\citep{ge2020self}. Recent works mainly focus on label generation, label refinery, the assistance of extra information, and optimization of representation. BUC~\citep{lin2019bottom} proposed a bottom-up clustering approach to generate pseudo labels. To reduce pseudo label noise, DCML~\citep{chen2020deep} selected credible training samples and MMT~\citep{ge2020mutual} proposed a mutual learning scheme for better pseudo labels. JVTC~\citep{li2020joint} and CycAs~\citep{wang2020cycas} explore temporal information to refine visual similarity. Contrastive learning with feature memory bank has been widely used in many works to learn more robust representation ~\citep{zheng2021group}~\citep{chen2021ice}.  SpCL~\citep{ge2020self} progressively generated more reliable clusters for the unified contrastive loss. Cluster Contrast~\citep{dai2021cluster} proposed to store feature vectors and compute contrast loss in the cluster level.
Although great success has been made, this line of methods usually leverage a single model to learn the knowledge that it already has, making it hard to learn sufficient capability due to the lack of diverse discrimination. To alleviate this problem, we propose PEG based on multi-model training where diversity of discrimination can be explored and exploited by the evolution of networks.

\subsection{Multiple Model Ensemble}
There is a considerable number of previous works on ensembles with neural networks. 
Explicit ensemble methods often train a series of base-level networks and average the predictions across them as the final result, which have low efficiency during both training and testing~\citep{hansen1990neural}~\citep{perrone1992networks}~\citep{krogh1994neural}~\citep{dietterich2000ensemble}~\citep{huang2017snapshot}~\citep{lakshminarayanan2017simple}. Recently, implicit ensemble methods are explored to tackle this problem. 
A typical approach ~\citep{srivastava2014dropout}, ~\citep{wan2013regularization}, ~\citep{huang2016deep}, ~\citep{singh2016swapout} generally create a series of networks with shared weights during training and then implicitly ensemble them at test time. Another approach ~\citep{shen2019meal} focuses on label refinery by distilling and transferring knowledge from a variety of trained networks to a single network for higher discrimination capability. 
However, these supervised methods cannot be directly used on unsupervised re-ID tasks, especially when the training set and the testing set share non-overlapping label space. 
On the other hand, existing methods accomplish the ensemble on all base-level networks while they ignore the problem that a very weak base-level network could drag down the overall performance when included.
Commonly, ``All'' is not the ``Best''. In this work, we propose a cooperative game in the selection phase of the framework to find and preserve the optimal combination of base-level networks using the unlabeled data and obtain progressive ensemble by an iterative population evolutionary gaming under unsupervised conditions.

\subsection{Algorithmic Game Theory} 
Machine learning methods with multi-agent game are proposed to address various tasks, such as image generation~\citep{goodfellow2014generative}, attacks and defenses for deep learning~\citep{yuan2019adversarial}, playing computer games~\citep{vinyals2019grandmaster}~\citep{penghybrid}, etc. SVM can be considered as a game between two agents where one agent challenges the other to find the best hyper-plane after providing the most difficult points for classification. Generative adversarial networks (GANs)~\citep{goodfellow2014generative} train two networks, the discriminator and the generator, against each other in order to generate images that can pass for real data. These methods are designed for non-cooperative games where agents have contrary rewards. However, in this work, the selection of networks is modeled as a multi-agent cooperative game, where rewards are global and shared by all agents. Although methods with cooperative games have been explored for reinforcement learning~\citep{penghybrid}, they can not be used for such a computer vision task.
Our approach consider the Best-response dynamics in cooperative game theory to solve a Nash equilibrium of model selection strategy.

\subsection{Unsupervised Evaluation Metrics of Models}
Metrics used in person re-ID always depend on samples with ground truth, such as mean Average Precision (mAP) and Cumulative Match Characteristic (CMC) curve, which are calculated between model prediction and the corresponding ground truth labels. However, these supervised metrics are not available during unsupervised learning when labels of data are unknown, therefore, they cannot be used as the criterion of the model selection in our PEG framework. 
On the other hand, several unsupervised evaluation metrics which require no data label have been designed to measure the performance of clustering algorithms as internal evaluation metrics ~\citep{4766909}~\citep{baker1975measuring}~\citep{hubert1976general}~\citep{maulik2002performance}~\citep{halkidi2002clustering}. 
For example, the silhouette coefficient ~\citep{rousseeuw1987silhouettes} estimates the average distance between each point in one cluster and points in the nearest neighboring cluster. The Dunn index ~\citep{dunn1973fuzzy} calculates the ratio of the minimum of inter-cluster distance to the maximum of intra-cluster distance between samples. 
{Nevertheless, these cluster validations cannot be directly used for the evaluation of re-ID models, for example, by the quality of clustering with their extracted features under the same clustering algorithm. That's because the distribution of feature clusters cannot measure the performance of models, especially in unsupervised settings. For instance, the metrics may estimate well clustering of features even when the model is poor but only is trained to overfit on its inaccurate labels.}
In this paper, we propose a cross-reference scatter which approximately measures a model’s discrimination capability by indirectly estimating its predicted pseudo-labels. It utilizes the pseudo-labels to train a reference network for a few iterations and then observes the cohesion and separation of its feature space to estimate the discrimination of the evaluated model. This method mines the latent visual relationships between image samples and so can approximately estimate models' discrimination on unlabeled data.

\subsection{Population-Based Evolutionary Training} 
Population-based evolution has been widely studied to solve real-valued optimization problems. For distance metric learning, a related task of re-ID, EDML~\citep{fukui2013evolutionary} and its variants ~\citep{kalintha2019kernelized}~\citep{ali2020reinforcement} were proposed to optimize a linear or non-linear transformation using differential evolution. However, these approaches cannot address the training of deep neural networks in re-ID due to the large scale of learnable parameters. 
Our approach is inspired by and built upon another line of Population Based Training (PBT) ~\citep{jaderberg2017population}, which is originally proposed for optimization of hyperparameters of networks. PBT trains a population of networks and performs periodically a process of exploiting and exploring, leading to automatic learning of the best configurations. It has been proved effective for a suite of challenging problems, including Atari and StarCraft II of reinforcement learning~\citep{vinyals2019grandmaster}~\citep{jaderberg2019human}, training Generative Adversarial Network (GAN)~\citep{jaderberg2017population} and data augmentation~\citep{ho2019population}. However, such a population-based training of networks has not been explored in unsupervised conditions, in which the criterion of network selection is difficult to determine. On the other hand, existing PBT approaches follow the principle of best individual selection, while our method selects and preserves optimal groups of networks that are more complementary. We additionally incorporate mutual learning within the population into the framework, leading to superior performance on the unsupervised re-ID.


\begin{figure*}[t]
\centering
\includegraphics[width=1.0\linewidth]{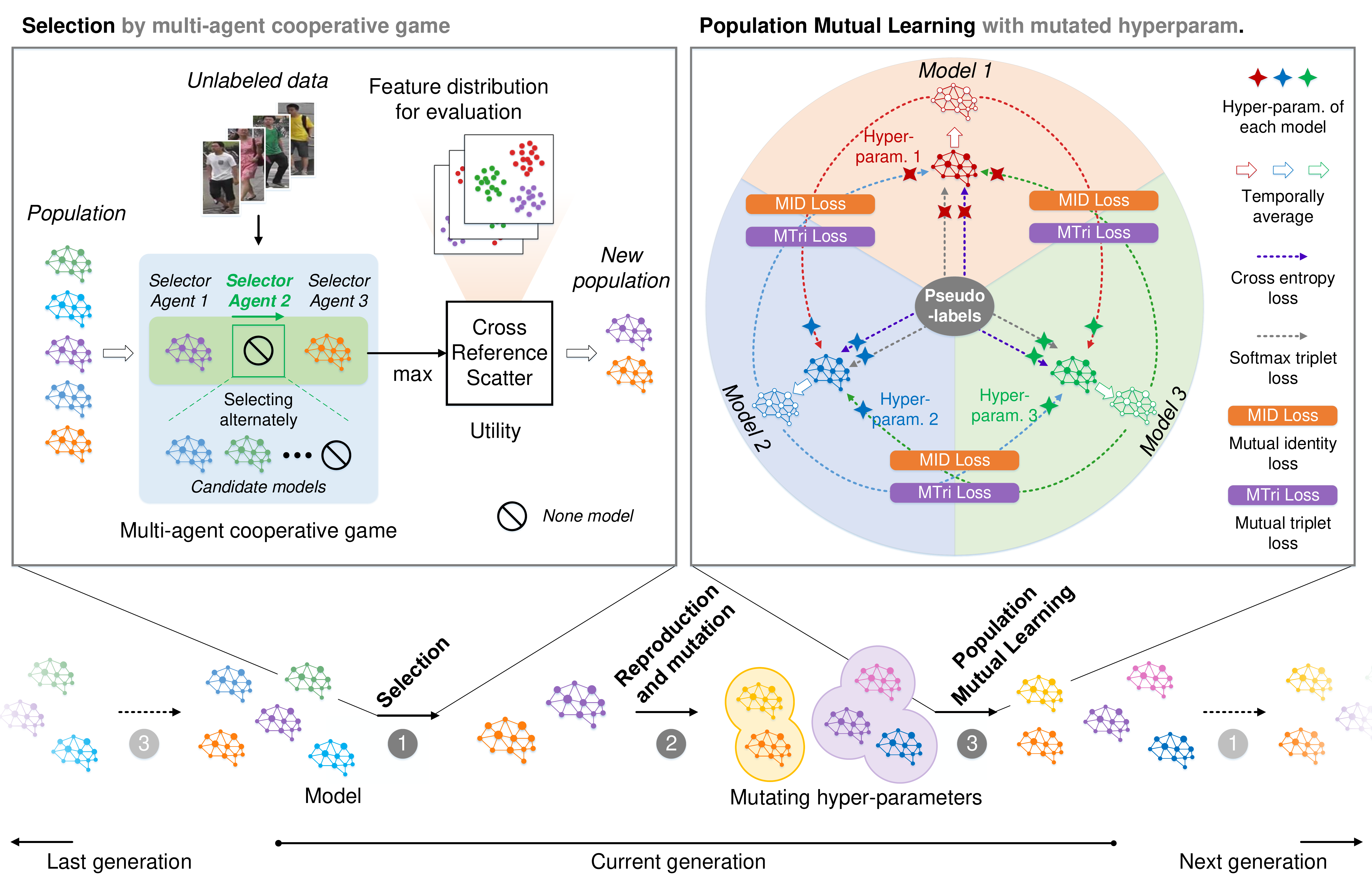}
\caption{\textbf{Population-based evolutionary gaming framework.}
PEG iteratively performs selection, reproduction\&mutation and population mutual learning to learn diverse and discriminative models under unsupervised conditions. In every generation, (1) selection preserves the optimal combination of models through a cooperative game with a set of selector agents to maximize the utility function (CRS). (2) Reproduction and mutation clone models and fluctuate their hyper-parameters to explore more diversity. (3) Population mutual learning trains models with mutated hyper-parameers by knowledge distillation from each other to enhance and assemble their discrimination.
}
\label{fig:overview}

\end{figure*}

\section{Methodology}
\subsection{Population-based Evolutionary Gaming}
\label{sec:peg}

Due to the lack of diversity of individual networks, sufficient discrimination for unsupervised person re-ID is difficult to achieve.
In contrast to previous works that use a single network for self-training, we propose a PEG that concurrently trains a diverse population of neural networks through an evolutionary game. In our formulation, the population $\mathcal{P}$ contains $K$ networks, each of which is denoted as $\mathcal{M}(\theta, \phi)$. $\theta$ is the learnable parameters, and $\phi$ is its hyper-parameters including the learning rate and loss ratios. The proposed training algorithm consists of three iterative phases, namely, selection to preserve adaptive networks, reproduction and mutation to learn more diversity, and population mutual learning to assemble knowledge, as illustrated in Fig. \ref{fig:overview}. The procedure of PEG is also described in \textbf{Algorithm} \ref{alg:algorithm-peg}.

\begin{algorithm*}[t]
\small
\caption{Population-based Evolutionary Gaming}
\label{alg:algorithm-peg}
\textbf{Input}: Unlabeled dataset $\{\mathbf{X}\}$. 

\textbf{Input}: Initial population $\mathcal{P}$ of $K$ models $\{\mathcal{M}^k\}$ with parameters $\{\theta^{k}\}$ and hyper-parameters $\{\phi^{k}\}$, $k=1,...,K$. 

\textbf{Output}: The inference model $\mathcal{M}(\theta)$.

\begin{algorithmic}[1]
\FOR{each generation}
\STATE // \textit{Selection} 
\STATE Select $L$ models from the population $\mathcal{P}$ by the cooperative game in Sec. \ref{sec:sel}, \\ \ $\{\mathcal{M}^l,l=1,...,L\}$=SELECTION($\mathcal{P}$, $L$).
\STATE Update the population $\mathcal{P} \leftarrow \{\mathcal{M}^l,l=1,...,L\}$.
\STATE //  \textit{Reproduction\&Mutation}
\FOR{each model $\mathcal{M}^l$}
\STATE Clone $H$ models of $\mathcal{M}^l$:  $\mathcal{M}^l_h(\theta^{l}_h,\phi^{l}_h)=\mathcal{M}^l(\theta^{l},\phi^{l}),h=1,...,H$.
\STATE Mutate the hyper-parameters of the cloned models: \\
$\phi^{l}_h  \sim \mathbf{U}((1-r)\phi^{l}_h, (1+r)\phi^{l}_h), h=1,...,H$.
\STATE Add the cloned model into the population, $\mathcal{P} \leftarrow \mathcal{P} + \{ \mathcal{M}^l_h(\theta^{l}_h,\phi^{l}_h) \}, h=1,...,H$
\ENDFOR
\STATE Update the population size $K \leftarrow L \times (H+1)$
\STATE //  \textit{Population mutual learning}
\STATE Optimize parameters of models in $\mathcal{P}$ by population mutual learning in Sec. \ref{sec:ml}:\\ $\{\theta^k\}\leftarrow$ PML($\mathbf{X},\{\theta^k\}$),$k=1,...,K$.
\ENDFOR
\STATE Select a model for inference: $\mathcal{M}(\theta)$=SELECTION($\mathcal{P}$, $1$)
\STATE \textbf{Return} The inference model $\mathcal{M}(\theta)$.
\end{algorithmic}
\end{algorithm*}

\subsubsection{Selection}
\label{sec:sel}
{ Since poor models may drag down the performance in multiple model training, we first propose a selection phase to preserve better models in PEG.}
Given a population $\mathcal{P}$ of $K$ neural network models $\{\mathcal{M}^1,...,\mathcal{M}^K\}$, selection aims to find an optimal subset of the population that is more adaptive to the given data, as shown in Fig. \ref{fig:overview}. Then, networks of the subset are preserved for later training, while other networks are abandoned to reduce the population size. The selection scheme is considered as a multi-agent cooperative game among $L$ selector agents characterized by $(\mathcal{A}_1,\mathcal{A}_2,...,\mathcal{A}_L, u)$, where $\mathcal{A}_l$ is the action space of agent $l$; and $u:\mathbf{A}\rightarrow \mathbb{R}$  denotes the utility function of the joint action $\mathbf{A} \in \mathcal{A}_1 \times \mathcal{A}_2 \times ...\times \mathcal{A}_L$. The action of each agent $\mathbf{a}_l \in \mathcal{A}_l$ is to select one neural network from the population $\mathcal{P}$, $\mathcal{A}_l = \{\mathcal{M}_1, ...,\mathcal{M}_K\}$. The number of agents is restricted due to the limitation of computational resources. In the cooperative game, agents pursue the same goal to maximize their team utility $u$. To maximize the discrimination and complementarity of the preserved networks, we define the utility function $u$ by the performance of the ensemble model. However, a model’s performance is difficult to estimate without labeled testing data. To address this problem, we design cross-reference scatter $J_{cr}$ to evaluate the ensemble model and consider it the formula of the utility function, $u(\mathbf{A})=J_{cr}(\vartheta(\mathbf{A}))$, where $\vartheta$ denotes the ensemble model produced by the networks currently selected by the agents. The detailed description of the cross-reference scatter will be provided in Section \ref{sec:crs}. Since there are approximately $K^L$ possible action combinations, a global optimal solution is impossible to derive by enumerating all the possibilities. Therefore, we turn to obtain a Nash equilibrium solution $\mathbf{\tilde{A}}=\{\mathbf{\tilde{a}}_l\}$, where each agent attempts to learn the best response to the other agents' actions:
\begin{equation}
u(\mathbf{a}'_l, \mathbf{\tilde{A}}_{-l}) \leq u(\mathbf{\tilde{A}}),
\label{eq:nash}
\end{equation}
where $\mathbf{a}'_l$ is any unilateral deviation, $\mathbf{\tilde{A}}_{-l}=\{\mathbf{\tilde{a}}_{l'}\}_{l'\neq l}$ and $-l$ represents all agents except $\mathbf{a}_{l}$.  
We solve Eq. \ref{eq:nash} via best-response dynamics. Each agent acts in a circular manner until it falls into a Nash equilibrium, where the action of each agent is the best response to the other agents. Below, we provide the detailed procedure of the cooperative selection game.
\begin{enumerate}[(1)]
\item Initialization of agent actions: randomly initialize $\mathbf{a}_l \in \mathcal{A}_l$ for $l=1,...,L$.
\item For agent $l$, solve the optimal action of agent $\mathbf{a}_l$ in response to the actions of the other $L-1$ agents. The objective for optimization of $\mathbf{a}_l$ can be formulated as,
\begin{equation}
\mathbf{a}_l^*= \mathop{\arg\max}_{\hat{\mathbf{a}}_l\in \mathcal{A}_l} u(\hat{\mathbf{a}}_l,\mathbf{a}_{-l}), \\
\label{eq:ops}
\end{equation}
where $\mathbf{a}_{-l}$ means all agents except $\mathbf{a}_{l}$.
\item Then update the joint action to $(\mathbf{a}_l^*, \mathbf{a}_{-l})$, as $\mathbf{a}_l^*$ is a best response to $\mathbf{a}_{-l}$. 
\begin{equation}
\mathbf{A} \leftarrow (\mathbf{a}_l^*, \mathbf{a}_{-l}).
\end{equation}
\item Repeat steps 2 to 3 for agent $l+1$.
\item If the joint action $\mathbf{A}$ has not changed in the last $L-1$ optimization rounds, the utility falls into Nash equilibrium, where every agent implements the best response to all other agents. In this case, we stop the optimization process, preserve the selected networks for the next generation, and abandon the other networks.
\end{enumerate}

{ \noindent\textbf{Cross-reference Scatter}}
\label{sec:crs}

A core issue is to estimate a model’s performance using unlabeled data in the selection phase of the proposed evolutionary game; however, such a measurement has not been explored. In this study, we propose a cross-reference scatter (CRS) for the approximate evaluation of re-ID models using unlabeled samples. Generally, the pseudo-labels predicted by better networks are more accurate, but the specific accuracy of the labels cannot be directly evaluated without the ground truth. However, models trained by more accurate pseudo-labels tend to achieve larger intra-cluster cohesion and inter-cluster separation in the feature space because incorrect labels will enforce models to separate samples of the same class or aggregate samples of different classes, which is difficult to accomplish. Therefore, it is reasonable to indirectly evaluate a network according to the feature cohesion and separation of a reference model that is trained by pseudo-labels which are predicted with the evaluated model.

\begin{figure}[t]
\centering
\includegraphics[width=1\linewidth]{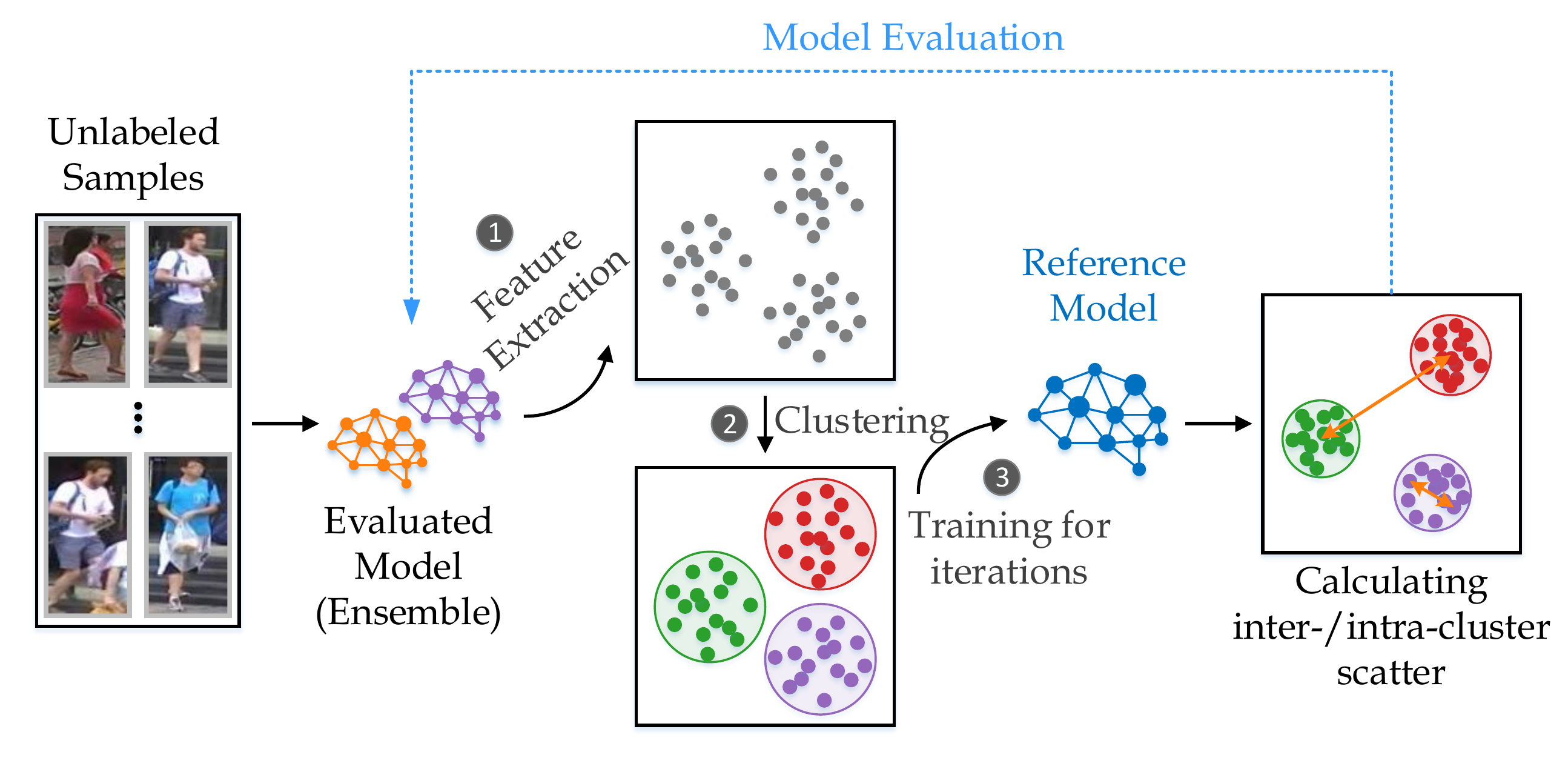}
\caption{Illustration of evaluation scheme of the proposed cross-reference scatter (CRS), which approximately measures a model’s discrimination capability by the inter-/intra-cluster scatter of a reference model that is briefly trained using pseudo-labels predicted by the evaluated model.}
\label{fig:crs}
\end{figure}

First, we introduce an inter-/intra-cluster scatter (ICS) to estimate the separation and cohesion of clusters in the feature space. Although existing metrics such as DBI \citep{4766909}, SC \citep{rousseeuw1987silhouettes} have been studied to estimate clustering, they usually pay more attention to the hard edge samples of clusters while ignoring the overall distribution, and thus are not applicable to measure re-ID models. Inspired by the objective of linear discriminant analysis, that is, to maximize the ratio of the between-class variance and the within-class variance, the inter-/intra-cluster scatter is defined as the ratio of the inter-cluster variance and intra-cluster variance in the clustered feature space. Given the set of images represented by feature vectors $\mathbf{f}(X|\Theta)$, where $\Theta$ denotes the parameters of the feature extractor network, we cluster all samples into $M$ groups as $\mathbb{C}$. 
We measure the cohesion of each cluster by the variance of features assigned to it.
The intra-cluster scatter of cluster $\mathbb{C}_i$ is defined as
\begin{equation}
\begin{aligned}
    S_{intra}^i(X|\Theta) = \sum_{x \in  \mathbb{C}_i}  [\mathbf{f}(x|\Theta) - \mu_i ]^T[\mathbf{f}(x|\Theta) - \mu_i],
\end{aligned}
\end{equation}
where $\mu_i=\sum_{x \in  \mathbb{C}_i} \mathbf{f}(x|\Theta)/n_i$ is the centroid of cluster $\mathbb{C}_i$ (with $n_i$ samples). 
Then, the intra-cluster scatter of all clusters is computed as 
\begin{equation}
\begin{aligned}
    S_{intra}(X|\Theta) = {\sum_{i=1}^{M} S_{intra}^i(X|\Theta)}.
\end{aligned}
\end{equation}
To measure the separation of feature clusters, the inter-cluster scatter is defined as the variance of cluster centroids
\begin{equation}
\begin{aligned}
    S_{inter}(X|\Theta) = \sum_{i=1}^{M} n_i [\mu_i - \mu]^T[\mu_i - \mu],
\end{aligned}
\end{equation}
where $\mu=\sum_{n=1}^{N} \textbf{f}(x_n|\Theta)/N$ is the center of the entire dataset. 
Considering both the separation and cohesion of feature clusters, the inter-/intra-cluster scatter $J(X|\Theta)$ is defined as the ratio of the inter-cluster scatter and intra-cluster scatter
\begin{equation}
\begin{aligned}
    J(X|\Theta) = {S_{inter}(X|\Theta)}/{S_{intra}(X|\Theta)}.
\end{aligned}
\label{eq:j}
\end{equation}
$J(X|\Theta)$ increases when the inter-cluster scatter is larger and the intra-cluster scatter is smaller, which entails larger separation and cohesion within feature clusters.

\begin{algorithm}[t]
\small
\caption{Cross Reference Scatter (CRS)}
\label{alg:algorithm3}
\textbf{Input}:  Unlabeled dataset $\{\mathbf{X}\}$. 

\textbf{Input}:  Evaluated model $\Theta$. 

\textbf{Input}:  Reference model $\theta^{ref}$.

\textbf{Output}: CRS of the evaluated model: $J_{cr}(\Theta)$.

\begin{algorithmic}[1]
\STATE Extract features on $\{\mathbf{X}\}$ by the evaluated model $\Theta$: $\mathbf{f}(\mathbf{X}|\Theta)$.
\STATE Generate pseudo-labels $\widetilde{Y}(\Theta)$ of $\mathbf{X}$ by clustering samples using $\mathbf{f}(\mathbf{X}|\Theta)$. 
\STATE Train the reference model $\theta^{ref}$ with $\{\mathbf{X},\widetilde{Y}(\Theta)\}$ for a fixed number of iterations by optimizing Eq. \ref{eq:id}, \ref{eq:tri}. 
\STATE Calculate ICS of the reference model $\theta^{ref}$ on $\{\mathbf{X}\}$: $J(\mathbf{X}|\theta^{ref})$ by Eq. \ref{eq:j}. 
\STATE $J_{cr}(\Theta) = J(\mathbf{X}|\theta^{ref})$.
\STATE \textbf{Return} CRS of the evaluated model: $J_{cr}(\Theta)$.
\end{algorithmic}
\end{algorithm}

Utilizing inter-/intra-cluster scatter (ICS), we attempt to evaluate a model by indirectly estimating its predicted pseudo-labels in a cross-reference (CR) manner. Given a network model with parameter $\Theta$ for evaluation, we first implement the model to extract the convolutional features of all samples $\mathbf{f}(X|\Theta)$. Then, minibatch k-means clustering is performed on $\mathbf{f}(X|\Theta)$ to classify all samples into $M$ different clusters.
After clustering, the produced cluster IDs are used as pseudo-labels $\widetilde{Y}(\Theta)$ for samples $X$. To estimate the accuracy of the predicted pseudo-labels $\widetilde{Y}(\Theta)$, we adopt them as supervision to train a reference model with parameters $\theta^{ref}$ by optimizing the cross-entropy loss with label smoothing and the softmax triplet loss for a certain number of iterations, and then measure the separation and cohesion of the reference model by computing the inter-/intra-cluster scatter $J(X|\theta^{ref})$ as cross reference scatter (CRS) $J_{cr}(\Theta)$. The value of CRS is then used for model evaluation, where a larger CRS indicates better discrimination capability of the evaluated model $\Theta$. The evaluation scheme is illustrated in Fig. \ref{fig:crs} and the detailed process is shown in \textbf{Algorithm} \ref{alg:algorithm3}.

{Importantly, we use the \textbf{k-means} clustering because the fair comparison of CRS among models requires the same cluster number during clustering. Specifically, since CRS is defined by the ratio of intra-cluster variance and inter-cluster variance, it is relative to the cluster numbers. For example, a larger cluster number may lead to a larger CRS value due to the smaller intra-cluster variances. And the cluster numbers by other clustering algorithms, i.e.  \textbf{DBSCAN}, with different evaluated feature models are likely to be different, making it unfair to compare their CRS for model selection.}

For fast and fair evaluation, we adopt a slight network, OSNet, with the same initial parameters as the reference model to evaluate different models. The number of training iterations of it is set to a small value from 500 to 1000 according to the number of samples.

\subsubsection{Reproduction and Mutation}
\label{sec:mutation}
Reproduction and mutation provide more diversity within the population by reproducing networks and mutating their hyper-parameters, including the learning rate and loss ratios after selection. In the reproduction and mutation phase, each network reproduces multiple descendants, one of which preserves the original hyper-parameters while the others apply a stochastic disturbance to their hyper-parameters to attempt to learn different information and increase the diversity of the population. Specifically, the mutated hyper-parameters $\phi'$ are sampled from a uniform distribution $\mathbf{U}((1-r)\phi, (1+r)\phi)$ that fluctuates within $r$ of the original value. 
The steps \textit{5-11} in \textbf{Algorithm} \ref{alg:algorithm-peg} summarize the process of reproduction and mutation.
Note that the mutation does not immediately change the weight parameters of neural networks. Changes occur to them when networks are trained by their mutated hyper-parameters in mutual learning.

\begin{algorithm*}[t]
\small
\caption{Population Mutual Learning (PML)}
\label{alg:algorithm2}
\textbf{Input}:  Unlabeled dataset $\{\mathbf{X}\}$. Population $\mathcal{P}$ of $K$ models parameterized by $\{\theta^{k}\}$, $k=1,...,K$.

\textbf{Output}: Updated neural network parameters $\{\theta^{k}\}$.

\begin{algorithmic}[1]
\FOR{each epoch}
\STATE Extract ensemble features on $\{\mathbf{X}\}$ by combinatorial model $\vartheta(\mathcal{P})$: $\mathbf{f}(\mathbf{X}|\vartheta(\mathcal{P})) = [ \mathbf{f}(\mathbf{X}|\theta^{1});...;\mathbf{f}(\mathbf{X}|\theta^{k})]$.
\STATE Generate pseudo-labels $\widetilde{Y}$ of $\mathbf{X}$ by clustering samples using $\mathbf{f}(\mathbf{X}|\vartheta(\mathcal{P}))$.
\FOR{each iteration $t$, mini-batch $\mathcal{B}\subset \mathbf{X}$ }
\STATE Randomly sample $S$ networks $\{\theta^{k_s}\} \subset \{\theta^{k}\}$, each indexed by $k_s$, $s=1,...,S$. 
\STATE Calculate soft-labels from temporally average model of each sampled network with $\{\Theta_{T}^{k_s}\}$: $\mathbf{p}(x_{i\in \mathcal{B}}|\Theta_{T}^{k_s})$, $\mathcal{P}_{i\in \mathcal{B}}({\Theta}_T^{k_s})$
\STATE Calculate output of each current model with $\{\theta^{k_s}\}$: $\mathbf{p}(x_{i\in \mathcal{B}}|\theta^{k_s})$, $\mathcal{P}_{i\in \mathcal{B}}({\theta}^{k_s})$.
\STATE Update parameters $\{\theta^{k_s}\}$ by optimizing Eq. \ref{eq:overall}.
\STATE Update temporally average model weights $\{\Theta_{t}^{k_s}\}$ following Eq.~\ref{eq:average}.
\ENDFOR
\ENDFOR
\STATE \textbf{Return} Networks parameters $\{\theta^k\}, k=1,...,K$.
\end{algorithmic}
\end{algorithm*}

\subsubsection{Population Mutual Learning}
\label{sec:ml}
After mutation, population mutual learning is performed among networks in the population $\mathcal{P}$ to access and assemble diverse discrimination capability using unlabeled data in an iteratively collaborative way, as shown in Fig. \ref{fig:overview}. Each network accomplishes learning from the whole population by means of its own hyper-parameters acquired from mutation. The learning scheme consists of a clustering-based pseudo-label prediction procedure and a mutual feature learning procedure. In each iterative epoch, pseudo-labels are first predicted for all samples via clustering and then utilized to fine-tune the networks of the population. 
In this phase, networks learn representations of images in two ways: from the shared pseudo-labels predicted by the whole population via clustering and from the output of other networks as soft labels via knowledge distillation. The procedure of this population mutual learning is described in \textbf{Algorithm} \ref{alg:algorithm2}.

\textbf{Pseudo-label prediction}. Pseudo-labels are predicted at the beginning of each iterative epoch. In order to predict reliable pseudo-labels, the framework utilizes all networks in the population $\{\mathcal{M}^1,...,\mathcal{M}^K\}$ jointly as a combinatorial model $\vartheta(\mathcal{P})$ to extract features for sample clustering. The clustering-based pseudo-label prediction procedure consists of three steps in total: (1) First, ensemble features of unlabeled samples $\{\mathbf{X}\}$ are obtained by concatenating features that are individually extracted by all networks, $\mathbf{f}(\mathbf{X}|\vartheta(\mathcal{P})) = [ \mathbf{f}(\mathbf{X}|\theta^{1});...;\mathbf{f}(\mathbf{X}|\theta^{k})]$; (2) Then, DBSCAN~\citep{DBLP:conf/kdd/EsterKSX96} clustering is performed on $\mathbf{f}(\mathbf{X}|\vartheta(\mathcal{P}))$to classify all unlabeled samples into $M$ different clusters. (3) The produced cluster IDs are used as pseudo-labels $\widetilde{Y}$ for the training samples $\mathbf{X}$. 
The steps 2 and 3 in \textbf{Algorithm} \ref{alg:algorithm2} summarize this clustering process.

{Different from CRS, we use DBSCAN here for model learning to generate more accurate pseudo-labels, since DBSCAN has been proven effective and efficient for a lot of clustering-based unsupervised person re-identification \cite{ge2020self} \cite{chen2021ice}. Compared with the k-means cluster algorithm, DBSCAN mines sample relations more accurately according to their density without setting the number of clusters and then helps learn more discriminative models. 
}

\textbf{Mutual feature learning}.
Utilizing the predicted pseudo-labels, the framework aims to organize networks within the population to learn from each other and enhance themselves in a mutual learning manner, as shown in Fig. \ref{fig:overview}. 
In each training iteration, the same batch of images with different random augmentations is first fed to all the networks in the population parameterized by $\{\theta^{k}\}$ to predict the classification confidence predictions $\{\mathbf{p}(x_{n}|\theta^{k})\}$ and feature representations $\{\mathbf{f}(x_{n}|\theta^{k})\}$. The classification confidence predictions are computed by a linear transformation of the feature representations followed by a softmax function. To transfer knowledge from one network to others, the outputs of each network serve as soft labels for training other networks. However, directly using the current predictions as soft labels to train each model decreases the independence of the model outputs, which might result in error amplification. To avoid this issue, the temporally averaged model \citep{tarvainen2017mean} of each network, which preserves more original knowledge, is used to generate reliable soft pseudo-labels for supervision. The parameters of the temporally averaged model of network $\theta^{k}$ at current iteration $t$ are denoted as $\Theta_t^{k}$, which is updated as
\begin{equation}
\label{eq:average}
\begin{aligned}
    {\Theta}_{t}^{k} = \alpha {\Theta}_{t-1}^{k} + (1-\alpha)\theta^{k},
\end{aligned}
\end{equation}
where $\alpha \in [0,1]$ is the scale factor, and the initial temporal average parameters are ${\Theta}_{0}^{k} = \theta^{k}$. 
For each network $\mathcal{M}^{k}$, three loss functions are computed as optimization objectives: mutual identity loss, mutual triplet loss and voting loss. 
The mutual identity loss \citep{zhang2018deep} of models learned by a certain network $\mathcal{M}^{e}$ is defined as the cross entropy between the ID prediction of the student network ${\mathcal{M}^k}$ and the teacher network $\mathcal{M}^{e}$
\begin{equation}
\begin{aligned}
    \mathcal{L}_{mid}^{k\leftarrow e} = - \frac{1}{N} \sum_{n=1}^{N} \mathbf{p}(x_{n}|\Theta^{e})^T \log \mathbf{p}(x_{n}|\theta^{k}). 
\end{aligned}
\label{eq:mid}
\end{equation}
The mutual triplet loss \citep{ge2020mutual} of models learned by a certain network $\mathcal{M}^{e}$ is defined as the binary cross entropy
\begin{equation}
\begin{aligned}
    \mathcal{L}_{mtri}^{k\leftarrow e} = -\frac{1}{N} \sum_{n=1}^{N} \bigg[ &\mathcal{P}_n({\Theta}^{e}) \log \mathcal{P}_n({\theta}^{k}) \\
    &+ (1-\mathcal{P}_n({\Theta}^{e})) \log (1-\mathcal{P}_n({\theta}^{k})) \bigg],
\end{aligned}
\label{eq:mtri}
\end{equation}
where $\mathcal{P}_n({\theta}^{k})$ denotes the softmax of the feature distance between negative sample pairs
\begin{equation}
\begin{aligned}
    \mathcal{P}_n({\theta}^{k}) = \frac{e^{\|\mathbf{f}(x_{n}|\theta^{k})-\mathbf{f}(x_{n-}|\theta^{k})\|}}{e^{\|\mathbf{f}(x_{n}|\theta^{k})-\mathbf{f}(x_{n+}|\theta^{k})\|}+e^{\|\mathbf{f}(x_{n}|\theta^{k})-\mathbf{f}(x_{n-}|\theta^{k})\|} },
\end{aligned}
\end{equation}
where $x_{n+}$ denotes the hardest positive sample of anchor $x_{n}$ according to the pseudo-labels $\widetilde{Y}$ and $x_{n-}$ denotes the hardest negative sample. $\|\cdot-\cdot\|$ denotes $L_2$ distance.

To learn stable and discriminative knowledge from the pseudo-labels obtained by clustering, we introduce voting loss, which consists of the classification loss and triplet loss. For each model ${\mathcal{M}^k}$, the classification loss is defined as the cross entropy with label smoothing \citep{szegedy2016rethinking}
\begin{equation}
\begin{aligned}
    \mathcal{L}_{id}^k = - \frac{1}{N} \sum_{n=1}^{N} \widetilde{\mathbf{q}}^T \log \mathbf{p}(x_{n}|\theta^{k}),
\end{aligned}
\label{eq:id}
\end{equation}
where $\widetilde{\mathbf{q}}$ is the smoothing label according to pseudo-labels $\widetilde{Y}$. Each element is calculated by
\begin{equation}
\begin{aligned}
    \widetilde{\mathbf{q}}_j = \begin{cases} 1-\varepsilon + \frac{\varepsilon}{M} & j=\widetilde{y}_{n}\\
     \frac{\varepsilon}{M} & j\neq \widetilde{y}_{n}
    \end{cases},
\end{aligned}
\end{equation}
The softmax triplet loss is defined as:
\begin{equation}
\begin{aligned}
    &\mathcal{L}_{tri}^k = \\
    &- \frac{1}{N} \sum_{n=1}^{N}  \log \frac{e^{\|\mathbf{f}(x_{n}|\theta^{k})-\mathbf{f}(x_{n-}|\theta^{k})\|}}{e^{\|\mathbf{f}(x_{n}|\theta^{k})-\mathbf{f}(x_{n+}|\theta^{k})\|}+e^{\|\mathbf{f}(x_{n}|\theta^{k})-\mathbf{f}(x_{n-}|\theta^{k})\|} }
\end{aligned}
\label{eq:tri}
\end{equation}
where $x_{n+}$ denotes the hardest positive sample of anchor $x_{n}$ according to the pseudo-labels and $x_{n-}$ denotes the hardest negative sample. The voting loss is defined by summarizing the classification loss and the triplet loss
\begin{equation}
\label{eq:vot}
\begin{aligned}
    \mathcal{L}_{vot}^k = w_{id} \mathcal{L}_{id}^k + w_{tri} \mathcal{L}_{tri}^k,
\end{aligned}
\end{equation}
where $w_{id}$ and $w_{tri}$ are the loss ratios.
For each model ${\mathcal{M}^k}$, the overall optimized objective is defined by 
\begin{equation}
\begin{aligned}
    \mathcal{L}^k = \frac{1}{K-1} ( w_{mid} \sum_{e\neq k}^{K} \mathcal{L}_{mid}^{k\leftarrow e} + w_{mtri} \sum_{e\neq k}^{K} \mathcal{L}_{mtri}^{k\leftarrow e} ) + \mathcal{L}_{vot}^k.
\end{aligned}
\label{eq:overall}
\end{equation}
Each model is trained by its own hyper-parameters $\phi=\{ \varepsilon, w_{id}, w_{tri}, w_{mid}, w_{mtri} \}$ to explore different information.
In addition, direct descendants of the same networks do not learn from each other in the mutual learning phase since they acquire similar knowledge. Note that training of a large-size population requires unaffordable computational resources. To address this problem, we use a random sampling strategy of networks. Specifically, for each batch of data, mutual learning is performed only on a randomly sampled subset of networks, as shown in step 5 in \textbf{Algorithm} \ref{alg:algorithm2}.

\subsection{Analysis of Escaping Capacity}
Here we analyze the escaping capacity from the local optimum of our approach. 
The optimization of our approach can be formulated into two interactive phases. The first one is to optimize the label assignment of samples according to feature models
\begin{equation}
\mathop{\arg\min}_{\widetilde{Y}} f_{a}(\widetilde{Y}, \Theta, X ),
\end{equation}
where $\widetilde{Y}$ is the assigned labels and $f_{a}$ is the objective loss function of the phase which is determined by the used clustering algorithm. $\Theta$ and $X$ denote the model parameters and input samples, respectively.
The second phase is to optimize the parameters of feature models according to the label assignment
\begin{equation}
\mathop{\arg\min}_{\Theta} f_{m}(\Theta, \widetilde{Y}, X |\phi).
\end{equation}
$f_{m}(\cdot|\phi)$ is loss function to train the models $\Theta$ as Eq. \ref{eq:overall}, and $\phi$ is the hyper-parameters.
The two optimization phases interact as a two-agent game and the \textit{local optimum} occurs when the game halts at a Nash equilibrium:
\begin{equation}
\begin{aligned}
&\exists~ (\widetilde{Y}^*, \Theta^*),\\
s.t. \\
&\widetilde{Y}^* = \mathop{\arg\min}_{\widetilde{Y}} f_{a}( \widetilde{Y}, \Theta^*, X),\\
&\Theta^* = \mathop{\arg\min}_{\Theta} f_{m}(\Theta,\widetilde{Y}^*, X |\phi).
\end{aligned}
\end{equation}
In our approach, function $f_{m}(\cdot|\phi)$ will be changed when the hyper-parameters $\phi$ are mutated to new values $\phi'$, leading to the shift of the local optimal model parameters $\Theta^*$. Therefore the Nash equilibrium between $\widetilde{Y}$ and $\Theta$ will be broken, and the local optimum at $(\widetilde{Y}^*, \Theta^*)$ will not exist exactly since the mutation changes the condition of the Nash equilibrium.

\section{Experiments}

\subsection{Datasets and Evaluation Metrics}

We evaluate the proposed method on three large-scale person re-identification benchmarks including Market-1501~~\citep{Zheng_2015_ICCV}, DukeMTMC-reID~~\citep{DBLP:conf/eccv/RistaniSZCT16}~\citep{Zheng_2017_ICCV} and MSMT17~\citep{Wei_2018_CVPR}.

\textbf{Market-1501:} 
This dataset contains 32,668 images of 1,501 identities from 6 disjoint cameras, among which 12,936 images from 751 identities form a training set, 19,732 images from 750 identities (plus a number of distractors) form a gallery set, and 3,368 images from 750 identities form a query set.

\textbf{DukeMTMC-reID:} This dataset is a subset of the DukeMTMC. It consists of 16,522 training images,
2,228 query images, and 17,661 gallery images of 1,812 identities captured using 8 cameras. Of the 1812 identities,
1,404 appear in at least two cameras and the rest (distractors) appear in a single camera.

{
\textbf{MSMT17}  contains 126,441 images of 4,101 IDs
captured from a 15-camera network. The training set has
32,621 images of 1,041 identities, and the testing
set has 93,820 images of 3,060 identities. During inference, 11,659 images are selected as query and the other 82,161 images are used as gallery from the testing set.
}

\textbf{Evaluation Metrics:} 
We use the Cumulative Matching Characteristic (CMC) curve and mean average precision (mAP) for performance evaluations and comparisons.

\subsection{Implementation Details}
\textbf{Model settings.} We adopt eight models with architectures of similar-weight parameters to initialize the population, including DenseNet-121~\citep{huang2017densely}, DenseNet-169, IBN-DenseNet-121~\citep{pan2018two}, IBN-DenseNet-169, Inception-v3~\citep{szegedy2016rethinking}, ResNet-50, IBN-ResNet-50a and IBN-ResNet-50b. All model are pretrained using ImageNet~\citep{DBLP:conf/cvpr/DengDSLL009}. In every model, the convolutional feature output by the last pooling layer is used for image representation. 

\noindent\textbf{PEG settings.}
The maximum size of networks in the selection phase $L$ is set to 3 for experiments. A lightweight OSNet~\citep{zhou2019omni} is used as the reference model of CRS for faster training. In addition, we conduct minibatch k-means clustering for CRS, and the number of clusters $M$ is set to 500 for Market-1501 and DukeMTMC-reID following MMT~\citep{ge2020mutual}. 
In the reproduction and mutation phase, each network reproduces 3 networks with mutation factor $r=0.5$. 
The whole population evolves for 3 generations in total. Our method is trained on 4 GPUs under PyTorch framework.
During testing, we use only one network which is selected by CRS in the population for feature representations.

\noindent\textbf{Training settings.}
In mutual learning, we calculate k-reciprocal Jaccard distance~\citep{Zhong_2017_CVPR} for clustering, where $k_1, k_2$ are set to 6 and 30, respectively. We set the minimum cluster samples to 4 and a distance threshold
to 0.6 for DBSCAN~\citep{DBLP:conf/kdd/EsterKSX96}.
During training, the input image is resized to $256\times128$, and traditional image augmentation is performed via random flipping and random erasing.
For each class from the training set, a mini-batch of 256 is sampled with $P$ = 16 randomly selected classes and $K$ = 16 randomly sampled images for computing the hard batch triplet loss. We use the Adam ~\citep{kingma2014adam} with weight decay 0.0005 to optimize parameters. In population mutual learning, the learning rate is fixed to 0.00035 for the overall 15 epochs. In each epoch, the temporal ensemble momentum $\alpha$ in Eq. \ref{eq:average} is set to 0.999.

\begin{table*}[t]
\caption{Comparison with person re-identification state-of-the-art methods on Market-1501 and DukeMTMC-reID datasets. ``*'' denotes the methods using extra temporal information. 
\textit{PEG(Full)} denotes the overall performance of our approach. For a fair comparison, \textit{PEG/ResNet50} is tested with the same ResNet50 backbone as most compared methods.
\textit{PEG+CCL} and \textit{PEG+ICE} denote training with ClusterContrast \citep{dai2021cluster} and ICE\citep{chen2021ice} as baselines, respectively. 
The performance of our approach is highlighted with bold fonts.
}
\begin{center}
\begin{tabular}{l||p{1.cm}<{\centering}p{1.cm}<{\centering}p{1.cm}<{\centering}p{1.cm}<{\centering}||p{1.cm}<{\centering}p{1.cm}<{\centering}p{1.cm}<{\centering}p{1.cm}<{\centering}}
    \hline\hline
    \multirow{2}{*}{Methods}  & \multicolumn{4}{c||}{\textbf{Market-1501}} & \multicolumn{4}{c}{\textbf{DukeMTMC-reID}} \\
    \cline{2-9}
    & mAP & R-1 & R-5 & R-10 & mAP & R-1 & R-5 & R-10  \\
    \hline\hline
    \multicolumn{9}{l}{\textbf{Unsupervised Domain Adaptation}}
     \\
    MMCL~\citep{wang2020unsupervised} & 60.4 & 84.4 & 92.8 & 95.0 & 51.4 & 72.4 & 82.9 & 85.0 \\
    JVTC~\citep{li2020joint} &  61.1 & 83.8 & 93.0 & 95.2 & 56.2 & 75.0 & 85.1 & 88.2 \\
    DG-Net++~\citep{zou2020joint} &  61.7 & 82.1 & 90.2 & 92.7 & 63.8 & 78.9 & 87.8 & 90.4 \\
    ECN++ ~\citep{zhong2020learning}  & 63.8 & 84.1 &92.8 & 95.4 &54.4 &74.0 & 83.7 & 87.4 \\
    AD-Cluster~\citep{Zhai_2020_CVPR}  & 68.3 & 86.7 & 94.4 & 96.5 & 54.1 & 72.6 & 82.5 & 85.5 \\
    MMT~\citep{ge2020mutual}  & 71.2 &87.7 &94.9 &96.9 &65.1 &78.0 &88.8 &92.5 \\
    DCML~\citep{chen2020deep} & 72.6 &87.9 &95.0 &96.7 &63.3 &79.1 &87.2 &89.4 \\
    MEB-Net~\citep{zhai2020multiple}  & {76.0} & {89.9} &   {96.0} & {97.5} &  {66.1} & {79.6} & {88.3} & {92.2} \\
    MetaCam-DSCE~\citep{yang2021joint} & 76.5 & 90.1 & -  & -  & 65.0 & 79.5  & -  & - \\
    SpCL~\citep{ge2020self} & 77.5 & 89.7 & 96.1 & 97.6 & - & -& -& - \\
    GLT~\citep{zheng2021group} & {79.5} & {92.2} & {96.5} & {97.8} & {69.2} & {82.0} & {90.2} & {92.8} \\
    \hline\hline
    \multicolumn{9}{l}{\textbf{Fully Unsupervised -- Linear Classifier Based}}
     \\
    LOMO~\citep{Liao_2015_CVPR}     & 8.0 & 27.2 & 41.6 & 49.1   & 4.8  & 12.3 & 21.3 & 26.6  \\
    Bow~\citep{Zheng_2015_ICCV}     & 14.8 & 35.8 & 52.4 & 60.3  & 8.3  & 17.1 & 28.8 & 34.9  \\
    UMDL~\citep{Peng_2016_CVPR}     & 12.4 & 34.5 & 52.6 & 59.6  & 7.3 & 18.5 & 31.4 & 37.6 \\
    BUC~\citep{lin2019bottom} & 29.6 &61.9 &73.5 &78.2 &22.1 &40.4 &52.5 &58.2 \\
    SSL~\citep{lin2020unsupervised} &37.8 &71.7 &83.8 &87.4 &28.6 &52.5 &63.5 &68.9 \\
    JVTC~\citep{li2020joint} & 41.8 &72.9 &84.2 &88.7 &42.2 &67.6 &78.0 &81.6 \\
    MMCL~\citep{wang2020unsupervised} &45.5 &80.3 &89.4 &92.3 &40.2 &65.2 &75.9 &80.0 \\
    MPRD~\citep{ji2021meta} &51.1& 83.0 &91.3 &93.6 & 43.7& 67.4 &78.7 &81.8 \\
    HCT~\citep{zeng2020hierarchical} &56.4 &80.0 &91.6 &95.2 &50.7 &69.6 &83.4 &87.4 \\
    *CycAs~\citep{wang2020cycas} & 64.8 &84.8 &- &- &60.1 &77.9 &- &- \\
    GCL~\citep{chen2021joint} &66.8 &87.3 &93.5 &95.5 &62.8 &82.9 &87.1 &88.5 \\
    *UGA~\citep{wu2019unsupervised} & 70.3 &87.2 &- &- &53.3 &75.0 &- &- \\
    IICS~\citep{xuan2021intra} & 72.1 &88.8 &95.3 &96.9 &59.1 &76.9 &86.1 &89.8 \\
    IN unsup.~\citep{fu2021unsupervised} &72.4 & 87.8 &-&-& 64.9 & 80.3 & -&-\\
    \textbf{PEG/ResNet50}  & \textbf{82.8} & \textbf{92.8} & 
                        \textbf{97.5} & \textbf{98.7} & 
                        \textbf{70.4} & \textbf{82.2} & 
                        \textbf{90.8} & \textbf{93.6} \\
    \textbf{PEG(Full)}  & \textbf{84.3} & \textbf{93.7} & \textbf{97.8} & \textbf{98.5}  & \textbf{71.9} & \textbf{83.8} & \textbf{91.2} & \textbf{93.5}\\
    \hline\hline
    \multicolumn{5}{l}{\textbf{Fully Unsupervised -- Memory Bank Based}} \\
    SpCL~\citep{ge2020self} & 73.1& 88.1 &95.1& 97.0 &65.3 &81.2 &{90.3} &92.2 \\
    OPLG~\citep{zheng2021online} & 78.1 & 91.1 & 96.4 & 97.7 & 65.6 & 79.8 & 88.6 & 91.6 \\
    ICE(agnostic)~\citep{chen2021ice} & {79.5} & {92.0} & {97.0} & {98.1} & 67.2 & {81.3} & {90.1} & {93.0} \\
    ClusterContrast~\citep{dai2021cluster} & {82.6} & {93.0} &   {97.0} & {98.1} &  72.8 & {85.7} & {92.0} & {93.5} \\
    \textbf{PEG+CCL/ResNet50}  & \textbf{83.3} & \textbf{93.4} & 
                        \textbf{97.3} & \textbf{98.4} & 
                        \textbf{74.4} & \textbf{84.6} & 
                        \textbf{92.1} & \textbf{94.0} \\
    \textbf{PEG+CCL(Full)}  & \textbf{87.1} & \textbf{94.6} & 
                        \textbf{98.0} & \textbf{98.8} & 
                        \textbf{76.8} & \textbf{86.4} & 
                        \textbf{93.1} & \textbf{95.0} \\
    \hline
    CAP~\citep{wang2020camera} &79.2 &91.4 &96.3 &97.7 &{67.3} &81.1 &89.3 &91.8 \\
    ICE(aware)~\citep{chen2021ice} & 82.3 &93.8 &97.6 &98.4& 69.9 &83.3 &91.5 &94.1 \\
    \textbf{PEG+ICE/ResNet50}  & \textbf{83.3} & \textbf{94.1} & 
                        \textbf{97.8} & \textbf{98.5} & 
                        \textbf{71.0} & \textbf{84.4} & 
                        \textbf{92.0} & \textbf{94.3} \\
    \textbf{PEG+ICE(Full)}  & \textbf{84.5} & \textbf{94.3} & 
                        \textbf{98.0} & \textbf{98.5} & 
                        \textbf{72.8} & \textbf{85.3} & 
                        \textbf{92.5} & \textbf{94.3} \\

\hline\hline
\end{tabular}
\end{center}
\label{table:staart}
\end{table*}

\begin{table*}[h]
\caption{Comparison with person re-identification state-of-the-art methods on MSMT17 dataset. ``*'' denotes the methods using extra temporal information. 
\textit{PEG(Full)} denotes the overall performance of our approach. For a fair comparison, \textit{PEG/ResNet50} is tested with the same ResNet50 backbone as most compared methods.
\textit{PEG+CCL} and \textit{PEG+ICE} denote training with ClusterContrast \citep{dai2021cluster} and ICE\citep{chen2021ice} as baselines, respectively. 
The performance of our approach is highlighted with bold fonts.
}
\footnotesize
\begin{center}
\begin{tabular}{l||p{1.1cm}<{\centering}p{1.1cm}<{\centering}p{1.1cm}<{\centering}p{1.1cm}<{\centering}}
    \hline\hline
    \multirow{2}{*}{Methods}  & \multicolumn{4}{c}{\textbf{MSMT17}} \\
    \cline{2-5}
    & mAP & R-1 & R-5 & R-10   \\
    \hline\hline
    \multicolumn{5}{l}{\textbf{Unsupervised Domain Adaptation}}
     \\
    ECN ~\citep{zhong2020learning}  & 10.2 & 30.2 & 41.5 & 46.8  \\
    MMT~\citep{ge2020mutual}  & 24.0 & 50.1 & 63.5 & 69.3  \\
    SpCL~\citep{ge2020self} & 26.8 & 53.7 & 65.0 & 69.8  \\
    \hline\hline
    \multicolumn{5}{l}{\textbf{Fully Unsupervised -- Linear Classifier Based}}
     \\
    MMCL~\citep{wang2020unsupervised} &11.2 &35.4 &44.8 &49.8 \\
    TAUDL~\citep{li2018unsupervised} & 12.5 & 28.4 & - & - \\
    UTAL~\citep{li2019unsupervised} & 13.1 & 31.4 & - & - \\
    IICS~~\citep{xuan2021intra} & 18.6 &45.7 &57.7 &62.8 \\
    *UGA ~\citep{wu2019unsupervised} & 21.7 & 49.5 &- & - \\
    *CycAs~\citep{wang2020cycas} & 26.7 & 50.1 &- &-  \\
    \textbf{PEG/ResNet50}  & \textbf{24.5} & \textbf{48.4} & 
                        \textbf{61.5} & \textbf{67.5}  \\  
    \textbf{PEG(Full)}  & \textbf{30.9} & \textbf{57.9} & \textbf{69.7} & \textbf{74.5} \\
    \hline\hline
    \multicolumn{5}{l}{\textbf{Fully Unsupervised -- Memory Bank Based}} \\
    SpCL~\citep{ge2020self} & 19.1& 42.3 &55.6& 61.2  \\
    ICE(agnostic)~\citep{chen2021ice} &29.8 &59.0& 71.7& 77.0\\
    ClusterContrast~\citep{dai2021cluster} & {27.6} & {56.0} &   {66.8} & {71.5}  \\
    \textbf{PEG+CCL/ResNet50}  & \textbf{33.4} & \textbf{61.3} & 
                        \textbf{73.4} & \textbf{77.8}  \\   
    \textbf{PEG+CCL(Full)}  & \textbf{41.8} & \textbf{69.1} & 
                        \textbf{79.5} & \textbf{82.9}  \\ 
    \hline
    CAP~\citep{wang2020camera} & 36.9& 67.4& 78.0 &81.4 \\
    ICE(aware)~\citep{chen2021ice}  &38.9 &70.2& 80.5& 84.4 \\
    \textbf{PEG+ICE/ResNet50}  & \textbf{42.1} & \textbf{72.0} & 
                        \textbf{82.0} & \textbf{85.4}  \\   
    \textbf{PEG+ICE(Full)}  & \textbf{44.9} & \textbf{73.9} & 
                        \textbf{83.2} & \textbf{86.3}  \\     
\hline\hline
\end{tabular}
\end{center}
\label{table:staartmsmt}
\end{table*}

\subsection{Comparison with State-of-the-Arts}

{
We compare PEG with state-of-the-art person re-ID methods in Table \ref{table:staart} and Table \ref{table:staartmsmt} on Market-1501, DukeMTMC-reID and MSMT17 datasets, respectively. 
The performance of our full approach is reported as \textit{PEG(Full)}.
In addition, we also evaluate PEG using the same backbone of ResNet50 as most of the other methods since backbones are important for feature learning. However, the backbones of models are automatically selected in our selection phase. To guarantee a model with ResNet50 is preserved in the population, we limit PEG to choose at least one ResNet50 network at every time of selection. The results tested by ResNet50 are reported as \textit{PEG/ResNet50}.

Previous unsupervised methods can be categorized into unsupervised domain adaptation (UDA) and fully unsupervised (FU) methods.  State-of-the-art UDA methods are first listed and compared in Table 1 and Table 2,  including MMCL~\citep{wang2020unsupervised}, JVTC~\citep{li2020joint}, DG-Net++~\citep{zou2020joint}, ECN++~\citep{zhong2020learning}, AD-Cluster~\citep{zhai2020ad}, MMT~\citep{ge2020mutual}, DCML~\citep{chen2020deep}, MEB-Net~\citep{zhai2020multiple}, MetaCam-DSCE~\citep{yang2021joint}, SpCL~\citep{ge2020self} and GLT~\citep{zheng2021group}. All these methods usually rely on an annotated source domain to provide basic discrimination and transfer it to the target domain. 
Without any identity annotation from source domains, our proposed PEG outperforms all of them on Market-1501, DukeMTMC-reID datasets, and most of them on MSMT17 dataset except SpCL. The results indicate the better capacity of PEG to explore the information of the unlabeled data by exploiting the diversity of multiple models. 
On the other hand, although other approaches have also been proposed to utilize multiple models, such as MMT and MEB-Net, our PEG still surpasses them by exploring and exploiting the diversity of multiple models through evolutionary gaming. With mutation to provide more diverse discrimination, it automatically finds and preserves the optimal combination of networks from the population in every generation and thus achieves better performance in the end.  

State-of-the-art fully unsupervised methods are then listed and compared in Table \ref{table:staart} and Table \ref{table:staartmsmt} 
including BUC~\citep{lin2019bottom}, SSL~\citep{lin2020unsupervised}, JVTC~\citep{li2020joint}, MMCL~\citep{wang2020unsupervised}, MPRD~\citep{ji2021meta}, HCT~\citep{zeng2020hierarchical}, CycAs~\citep{wang2020cycas}, GCL~\citep{chen2021joint}, UGA~\citep{wu2019unsupervised}, IICS~\citep{xuan2021intra}, IN unsup.~\citep{fu2021unsupervised}, SpCL~\citep{ge2020self}, OPLG~\citep{zheng2021online}, CAP~\citep{wang2020camera}, ICE~\citep{chen2021ice} and ClusterContrast~\citep{dai2021cluster}. Especially, ICE (aware) denotes using extra camera information, and ICE (agnostic) denotes not using it. The compared approaches mainly rely on the pseudo-label discovery of single networks. 
{Among them, methods tagged by ``*'' denote that elaborate extra temporal information is additionally used to improve the discrimination, such as CycAs and UGA, while our approach only considers person appearance similarity. }
The performance of these methods is provided just for reference since it is not our point to explore the extra temporal information, and our method does not use any of them. 
The fully unsupervised methods are separated into two groups, including linear classifier based methods and memory bank based methods:

{

\textbf{(1) Comparison with linear classifier based methods.} 
For the linear classifier based methods, our approach with ResNet50 achieves better performance than most of them only except CycAs and UGA on the MSMT17 dataset, as shown in Table \ref{table:staart}, \ref{table:staartmsmt}. Different from the other two datasets, CycAs and UGA with extra temporal information achieves better performance on MSMT17 because images in the dataset are more diverse and harder to cluster accurately, making the elaborate extra temporal information particularly important. Nevertheless, these methods still suffer from the lack of diversity in single model training, which prevented them from maximizing their discrimination under unsupervised conditions.
The superior performance of PEG can be attributed to the multiple model training, which improves the networks’ discriminative capability by mutual learning among diverse networks. And it can also be attributed to the selection of PEG, which preserves the more discriminative models in every generation and achieves the better performance of them. 
In addition, our full approach PEG(Full) further improves the re-ID performance by automatically selecting better architectures.

\textbf{(2) Comparison with memory bank based methods.} 
Memory banks \citep{ge2020self} were employed in many recent methods \citep{dai2021cluster} to replace the linear classifier before the softmax cross-entropy loss function to improve unsupervised re-ID performance.
Specifically, memory bank based methods can be further categorized into two groups including i) ClusterContrast, SpCL, and OPLG that learn general memories for all cameras, and ii) ICE and CAP that design specific memories for each camera. 
Since our research mainly focuses on the problem of training multiple models, it is independent of these methods for training single models. And they are not contradictive with our main contribution and are compatible with our method. To verify this, we additionally report our performance on the two typical stronger baselines of ClusterContrast as \textit{PEG+CCL}, and ICE as \textit{PEG+ICE} respectively in Table \ref{table:staart}, \ref{table:staartmsmt}.

For the camera-general memory bank based methods, \textit{PEG+CCL / ResNet50} surpasses most of other state-of-the-art methods only except ClusterContrast for Rank-1 accuracy on DukeMTMC-reID.
For ClusterContrast, the relatively poor Rank-1 on DukeMTMC-reID dataset shows its weakness for some hard negative samples which were mistakenly identified as the same persons, because the soft mutual losses in mutual learning lack the certainty of labels and may not learn strong capability to separate hard negative samples.
However, robust improvement of PEG is mainly shown by other metrics, especially the higher mAP on all benchmarks, indicating that PEG deals better with those hard positive samples, which is more important for the practical application of security. 
Furthermore, our full approach of PEG + CCL (Full) produces a new state-of-the-art performance on Market1501 and DukeMTMC-reID.
The better performance can be attributed to the fact that the diverse population provides more reliable supervision for each other. The improved results also demonstrate that our evolution gaming approach is easily combined with different loss functions and can be further improved by more effective losses.

For the camera-specific memory bank based methods, \textit{PEG+ICE / ResNet50} outperforms all the compared methods on the three datasets and produces a new state-of-the-art performance on MSMT17 dataset. The superior performance to the PEG+CCL on MSMT17 can be attributed to that camera-specific memories alleviate the strong camera variance in the dataset which has 15 cameras. However, camera-specific memories are complementary with our PEG framework and can be further improved for better performance.

}

}

\begin{table*}[t]
\caption{Ablation studies of PEG using eight initial networks under unsupervised conditions. \textit{Single model baseline} denotes the best performance of single model training using self-improvement mechanism. \textit{Multi-model} means that all models are used for pseudo label prediction and every model is then trained individually. \textit{PML} denotes population mutual learning. \textit{Sel.}, \textit{Rep.} and \textit{Mut.} denote selection, reproduction and mutation in PEG framework, respectively. { \textit{Single architecture + PEG} denotes the population is initialized with a single model.}}
\begin{center}
\begin{tabular}{l|p{0.9cm}<{\centering}p{0.9cm}<{\centering}p{0.9cm}<{\centering}p{0.9cm}<{\centering}|p{0.9cm}<{\centering}p{0.9cm}<{\centering}p{1.0cm}<{\centering}p{1.0cm}<{\centering}}
    \hline\hline
    \multirow{2}{*}{Methods} & \multicolumn{4}{c|}{Market-1501} & \multicolumn{4}{c}{DukeMTMC-reID} \\
    \cline{2-9}
    & mAP & R-1 & R-5 & R-10 & mAP  & R-1 & R-5 & R-10 \\
    \hline\hline
    Single model baseline  & 69.6 & 84.9 & 92.9 & 94.9  & 55.9 & 72.8 & 81.9 & 85.4  \\
    Multi-model  & 76.4 & 89.7 & 95.6 & 97.1 & 63.4 & 79.0 & 87.0 & 90.0  \\
    Multi-model + PML  & 78.5 & 90.4 & 96.1 & 97.6 & 66.1 & 80.3 & 88.3 & 91.2  \\
    Multi-model + PML + Sel.    &  79.2 & 90.9  & 96.3  & 97.5  & 66.9 &  80.8 &  88.8  &  91.6 \\
    
    Multi-model + PML + Sel. + Rep.\&Mut.  & 84.3 & 93.7 & 97.8 & 98.5  & 71.9 & 83.8 & 91.2 & 93.5  \\
    {Single architecture + PEG}   & 80.1 & 90.9 & 96.3 & 97.4  &  69.4&82.1  & 90.3   &92.9 \\
    \hline\hline
\end{tabular}
\end{center}
\label{table:ablation_peg}
\end{table*}

\begin{table*}[t]
\caption{Ablation studies of PEG on a stronger baseline of cluster contrast loss (CCL).  \textit{CCL-Single} denotes the baseline using the model of IBN-ResNet50. \textit{CCL-Multi} means that all models are used for pseudo label prediction and every model is then trained individually. \textit{Sel.} denotes selection, and  \textit{PEG} denote our full method.}
\begin{center}
\begin{tabular}{l|p{0.9cm}<{\centering}p{0.9cm}<{\centering}p{0.9cm}<{\centering}p{0.9cm}<{\centering}|p{0.9cm}<{\centering}p{0.9cm}<{\centering}p{1.0cm}<{\centering}p{1.0cm}<{\centering}}
    \hline\hline
    \multirow{2}{*}{Methods} & \multicolumn{4}{c|}{Market-1501} & \multicolumn{4}{c}{DukeMTMC-reID} \\
    \cline{2-9}
    & mAP & R-1 & R-5 & R-10 & mAP  & R-1 & R-5 & R-10 \\
    \hline\hline
    CCL-Single   & 83.3 & 92.6 & 96.8 & 97.9  & 74.4 & 86.2 & 92.1  &  93.9  \\
    CCL-Multi  & 79.4 & 91.2 & 96.8 & 97.8 & 68.9 & 81.6 & 89.9 & 91.7  \\
    CCL-Multi + Sel.    &  84.3 &  92.9 & 97.1  & 98.2  & 74.9 &  86.3 & 92.4  & 94.1  \\
    CCL-Multi + PEG  & 87.1 & 94.6 & 98.0 & 98.8  & 76.8 & 86.4 & 93.1 & 95.0  \\
    \hline\hline
\end{tabular}
\end{center}
\label{table:ablation_pegccl}
\end{table*}

\begin{table*}[h]
\caption{Ablation studies of components of population mutual learning (PML) on selected models without reproduction and mutation: {\textit{Supervised Upper Bound} - Deep models trained using the labelled training images. \textit{Single Model} - evaluation using the best single model.  \textit{Ensemble Feature} - evaluation using feature ensemble among multiple networks. }
\textit{Baseline Ensemble} - Models jointly trained by shared pseudo-labels but without mutual learning. $\mathcal{L}_{vot}$ (Eq.~\ref{eq:vot}), $\Theta_T$ (Eq.~\ref{eq:average}), $\mathcal{L}_{mid}$ (Eq.~\ref{eq:mid}) and  $\mathcal{L}_{mtri}$ (Eq.~\ref{eq:mtri}) are described in Sec. \ref{sec:ml}.}
\begin{center}
\footnotesize
\begin{tabular}{l|cccc|cccc}
    \hline\hline
    \multirow{2}{*}{Methods} & \multicolumn{4}{c|}{Market-1501} & \multicolumn{4}{c}{DukeMTMC-reID} \\
    \cline{2-9}
    & mAP & R-1 & R-5 & R-10 & mAP & R-1 & R-5 & R-10 \\
    \hline\hline
    \multicolumn{5}{l}{\textbf{Supervised Upper Bound}} \\
    {Single Model} & {84.4 }& {94.1 }& {97.9 }& {98.8 } &{ 71.2} & {85.5 }& {92.5 }& {94.3 } \\
    {Ensemble Feature} & {86.4 }& {94.9 }& {98.0 }& {98.9 } &{ 77.7} &{ 88.9} & {94.5 }& {95.6 } \\
    \hline
    Baseline Ensemble (Only $\mathcal{L}_{vot}$)  & 76.6 & 89.1 & 95.7 & 97.2 & 63.2 & 77.2 & 87.5 & 91.0  \\
    PML w/o $\Theta_T$            &  73.3 & 87.9  & 95.3  & 97.1  & 62.6 & 77.5  & 87.0  & 90.1  \\
    PML w/o $\mathcal{L}_{mid}$    &  77.2 & 90.2  & 95.9  & 97.4  & 65.1 & 79.2  & 88.5  & 91.2   \\
    PML w/o $\mathcal{L}_{mtri}$    & 77.0  & 89.6  & 95.8  & 97.4 & 65.3 & 79.3  & 88.8  & 91.2   \\
    \hline
    PML  &  79.2 & 90.9  & 96.3  & 97.5  & 66.9 &  80.8 &  88.8  &  91.6 \\
    \hline\hline
\end{tabular}
\end{center}
\label{table:ablation}
\end{table*}

\subsection{Ablation Study}
\subsubsection{Evaluation of Components}
\label{sec:ab-components}
Detailed ablation studies are performed to evaluate the components of PEG as shown in Table \ref{table:ablation_peg}.

\textbf{Effectiveness of multiple model training.} Multiple model training usually achieves better performance than single model training because of the complementary discrimination of different models. In this section, we first introduce a baseline multi-model ensemble without mutual learning for comparison that only uses voting loss in Eq. \ref{eq:vot} to train networks jointly, denoted as \textit{Multi-model}. With eight networks used for ensemble, pseudo-labels are predicted by concatenating the features outputted from all networks and then used to supervise the training of each network individually by optimizing the voting loss. We also report the result of the \textit{single model baseline} using the best architecture, ResNet50-IBNa. As shown in Table \ref{table:ablation_peg}, \textit{Multi-model} outperforms the single model training by large margins, indicating that more accurate pseudo-labels can be predicted using multiple models. 

\textbf{Effectiveness of population mutual learning.} Population mutual learning conducts knowledge distillation among all base models for the better ensemble. Compared with the baseline ensemble as \textit{Multi-model}, models achieve better performance with mutual learning among themselves, as \textit{Multi-model + PML} in Table \ref{table:ablation_peg}. For example, the mAP is improved by 2.1\% and 2.7\% on Market-1501 and DukeMTMC-reID, respectively. The improvement can be attributed to that models additionally learn the distribution predicted by other models which contain more discriminative information.

In addition, more detailed ablation studies are performed to evaluate the components of mutual learning as shown in Table~\ref{table:ablation}. In this experiment, three networks (DenseNet-121-IBNa, DenseNet-169-IBNa, and ResNet-50-IBNa) are trained concurrently.
We first validate the temporally average model by removing it, denoted as \textit{PML w/o $\Theta_T$}. For this experiment, we directly use the prediction of the current networks parameterized by $\theta_T$ instead of the temporally average networks with parameters $\Theta_T$ as soft labels. As Table. \ref{table:ablation} shows, distinct drops are observed, indicating that networks tend to degenerate to be homogeneous without using temporally average models, which substantially decreases the learning capability.
Then we evaluate the mutual loss in Sec. \ref{sec:ml} from two aspects: the mutual identity loss and the mutual triplet loss. The former is denoted as \textit{PML w/o $\mathcal{L}_{mid}$}. Results show that mAP drops from 79.2\% to 77.2\% on Market-1501 dataset and from 66.9\% to 65.1\% on DukeMTMC-reID dataset. Similar drops can also be observed when studying the mutual triplet loss, which is denoted as \textit{PML w/o $\mathcal{L}_{mtri}$}. The effectiveness of the mutual learning, including both two mutual losses, can be largely attributed to that it enhances the discrimination capability of all networks. Overall, the performance of the mutual learning ensemble largely outperforms the baseline ensemble.
{
We also compare the mutual learning ensemble with two supervised upper bounds, which are trained using ground truths. The \textit{Single Model} denotes evaluation using the best single model, and the \textit{Ensemble Feature} denotes evaluation using feature ensemble among multiple networks. Our mutual learning ensemble is relatively close to them with evaluation using a single model. }

\textbf{Effectiveness of selection.} Selection phase in PEG finds and preserves an optimal subset of base networks for better multi-model training. The experiment with mutual learning and selection is denoted as \textit{Multi-model + PML + Sel.} in Table \ref{table:ablation_peg}. For this experiment, the selection is performed to preserve a combination of 3 networks from all 8 networks using the cooperative game in Section \ref{sec:sel}, then the preserved models are trained by mutual learning. Experimental results show that the selection phase improves the performance of \textit{Multi-model + PML} even using fewer models. The superior performance indicates that some models may be redundant and cannot provide more discrimination but require more computation during training. However, the selection effectively preserves better models with the proposed cooperative gaming while abandoning weak models that could even degrade the overall discrimination capability of the whole ensemble. Without those weaker models, models will achieve better discrimination from more reliable and efficient mutual learning.

\textbf{Effectiveness of reproduction and mutation.} Reproduction and mutation drive the PEG framework to train more diverse models by mutating their hyper-parameters, which is the key to the exploration of model diversity in our evolution process. With this component, PEG achieves the best performance in Table \ref{table:ablation_peg} as \textit{Multi-model + Sel. + Rep.$\&$Mut.}. The effectiveness of reproduction and mutation can be attributed to the exploration of training more diverse models with selection preserving the better ones of them after mutual learning. Beneficial from the iteration of reproduction, mutation, and selection, PEG keeps exploring and exploiting diverse and discriminative capacity for better re-ID models.
{
In addition, multiple network ensemble with different architectures is also used to exploit their diversity. To further validate the effectiveness for the diversity of reproduction and mutation, we evaluate PEG with only a single model to initialize the population with reproduction and mutation, as \textit{Single architecture + PEG} in Table \ref{table:ablation_peg}. Compared with the single model baseline, the experiment improves the accuracy by large margins, and it also outperforms the multi-model without reputation and mutation. The results demonstrate that reputation and mutation are more important for exploring diversity. Moreover, our full approach of PEG with both multiple architectures and reputation\&mutation performs the best, demonstrating that the diversities from the two components are complementary. 
}

{
\textbf{Generalization Analysis.}
To validate the generalization of our approach with different baseline training methods, ablation studies on a stronger baseline of cluster contrast loss are evaluated as shown in Table \ref{table:ablation_pegccl}. Compared with the single model, it performs not good when directly using multiple models for pseudo-label prediction, denoted as \textit{CCL-Multi}. The distinct degradation of performance indicates that the weak models make a negative impact on such a stronger baseline. The models converge quickly to the inaccurate pseudo label partially predicted by the weak models and can no longer be improved. However, the performance of the multi-model is largely improved using the selection before training. It demonstrates that the selection still preserves better models effectively and abandons the weak models which are harmful to the ensemble. Furthermore, \textit{CCL-Multi + PEG} produces the best performance on both datasets, validating the effectiveness of the mutation and reproduction. The superior results show that our PEG is effective and generalizable for different baseline methods.   
}

\begin{figure*}[t]
\centering
\includegraphics[width=1.0\linewidth]{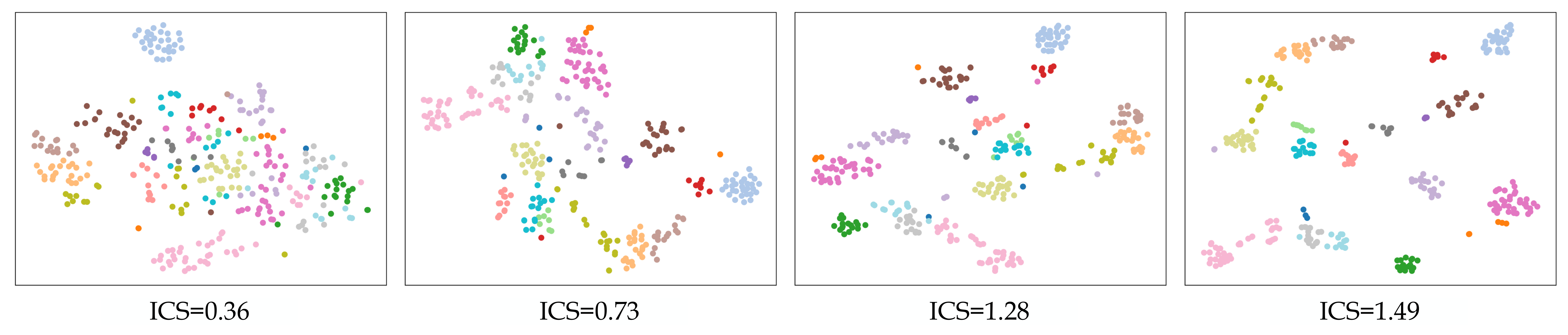}
\caption{Illustration of feature distribution with different Inter-/intra-cluster scatters (ICS). A larger ICS means larger cohesion within feature clusters and larger separation across feature clusters. Best viewed in color.}
\label{fig:scatters}
\end{figure*}

\begin{figure}[t]
\centering
\includegraphics[width=0.8\linewidth]{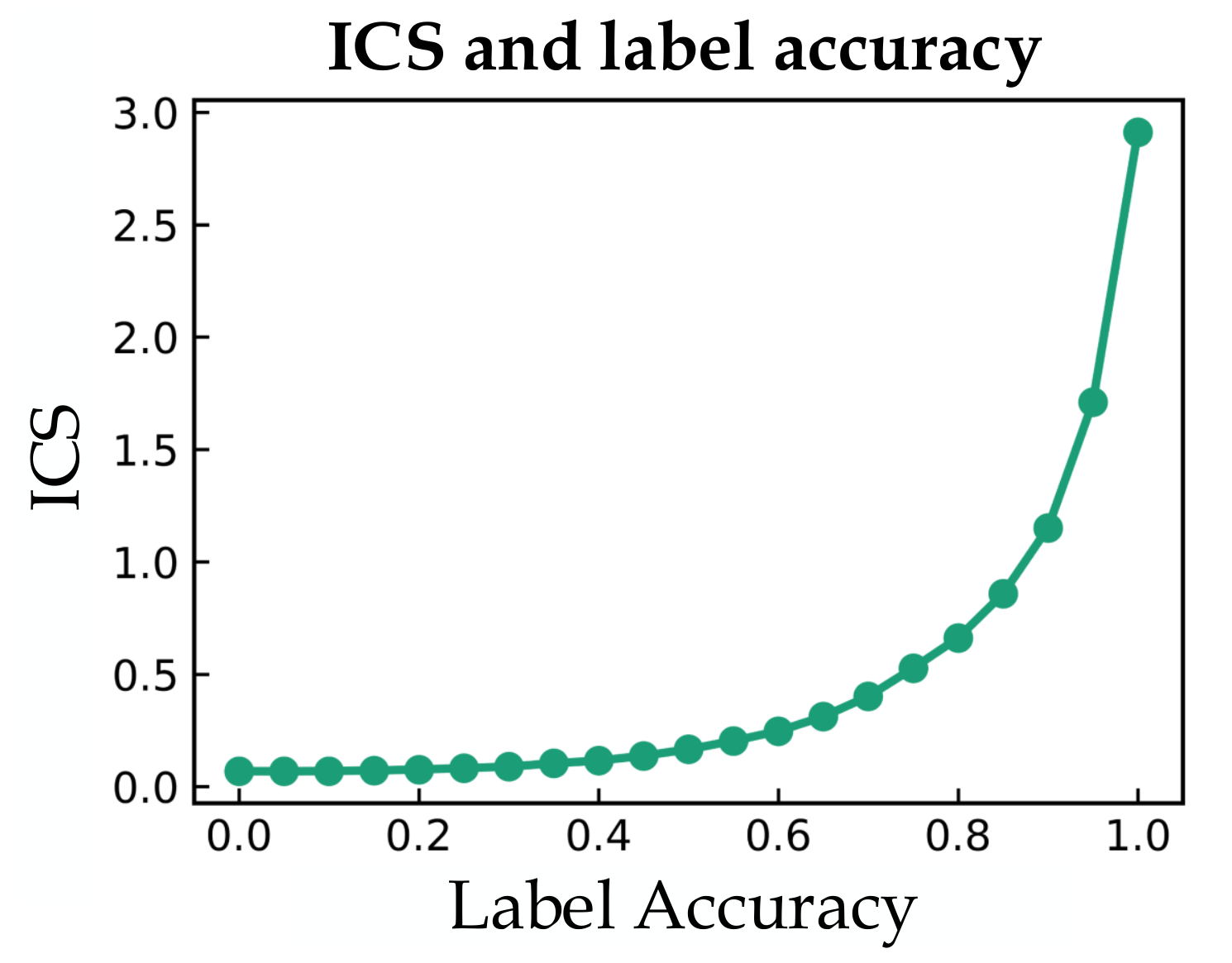}
\caption{The positive correlation between inter-/intra-cluster scatter (ICS) and label accuracy indicates that more accurate labels usually lead to larger ICS, which means larger cohesion within feature clusters and larger separation across feature clusters during model training.}
\label{fig:lacc}
\end{figure}

\begin{figure*}[t]
\centering
\includegraphics[width=1.0\linewidth]{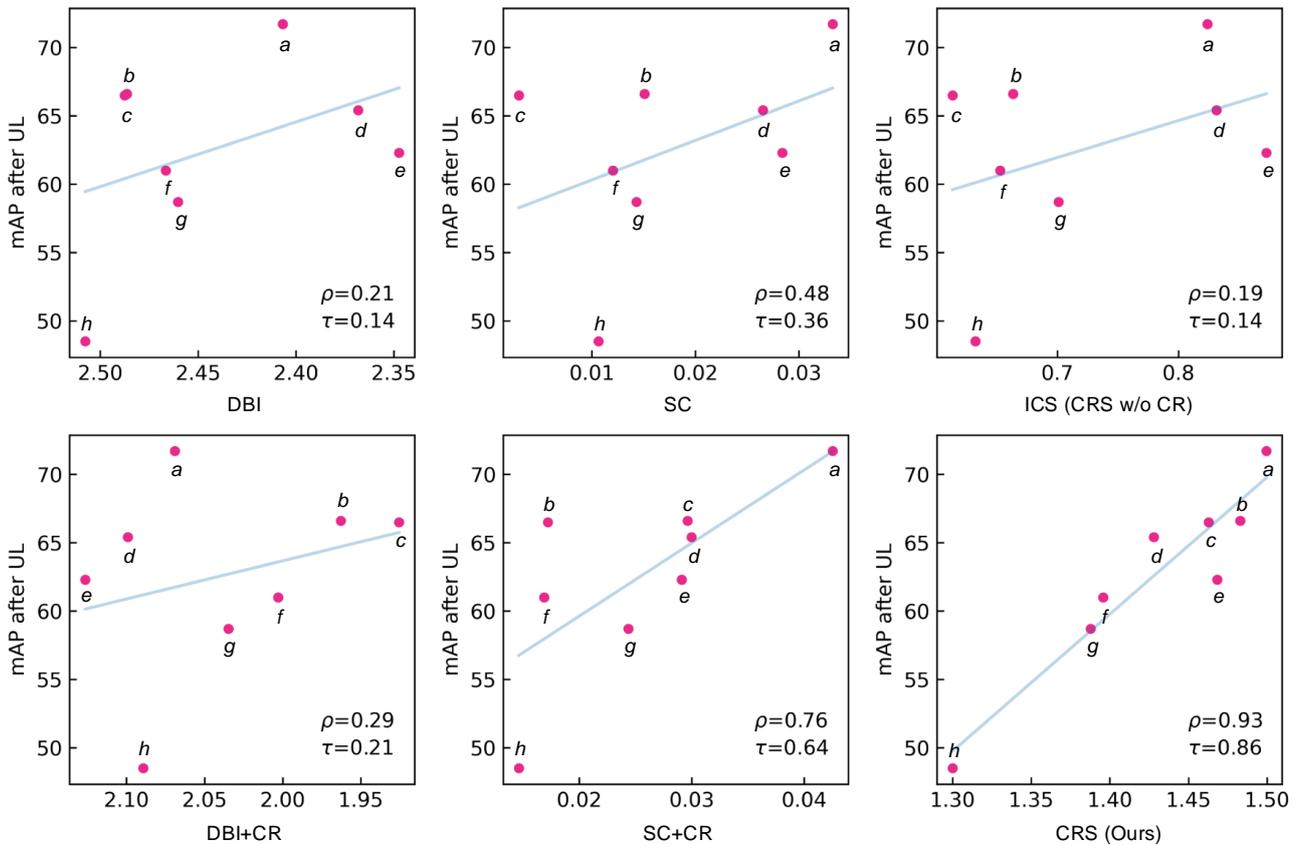}
\caption{Comparison with different unsupervised measures on the correlation between the measures and re-ID performance (mAP after unsupervised learning) over different models (point \textit{a-h}) on Market-1501 dataset. For each measure, we use Spearman's Rank Correlation ($\rho$) \citep{spearman1961proof} and Kendall's Rank Correlation ($\tau$) \citep{kendall1938new} to measure the correlation between the metric and mAP values. A higher absolute value of $\rho$ (or $\tau$) indicates a stronger correlation. Our proposed CRS shows a stronger correlation, indicating that it better reflects the performance of re-ID models.}
\label{fig:scatters_arch}
\end{figure*}

\begin{figure}[t]
\centering
\includegraphics[width=1.0\linewidth]{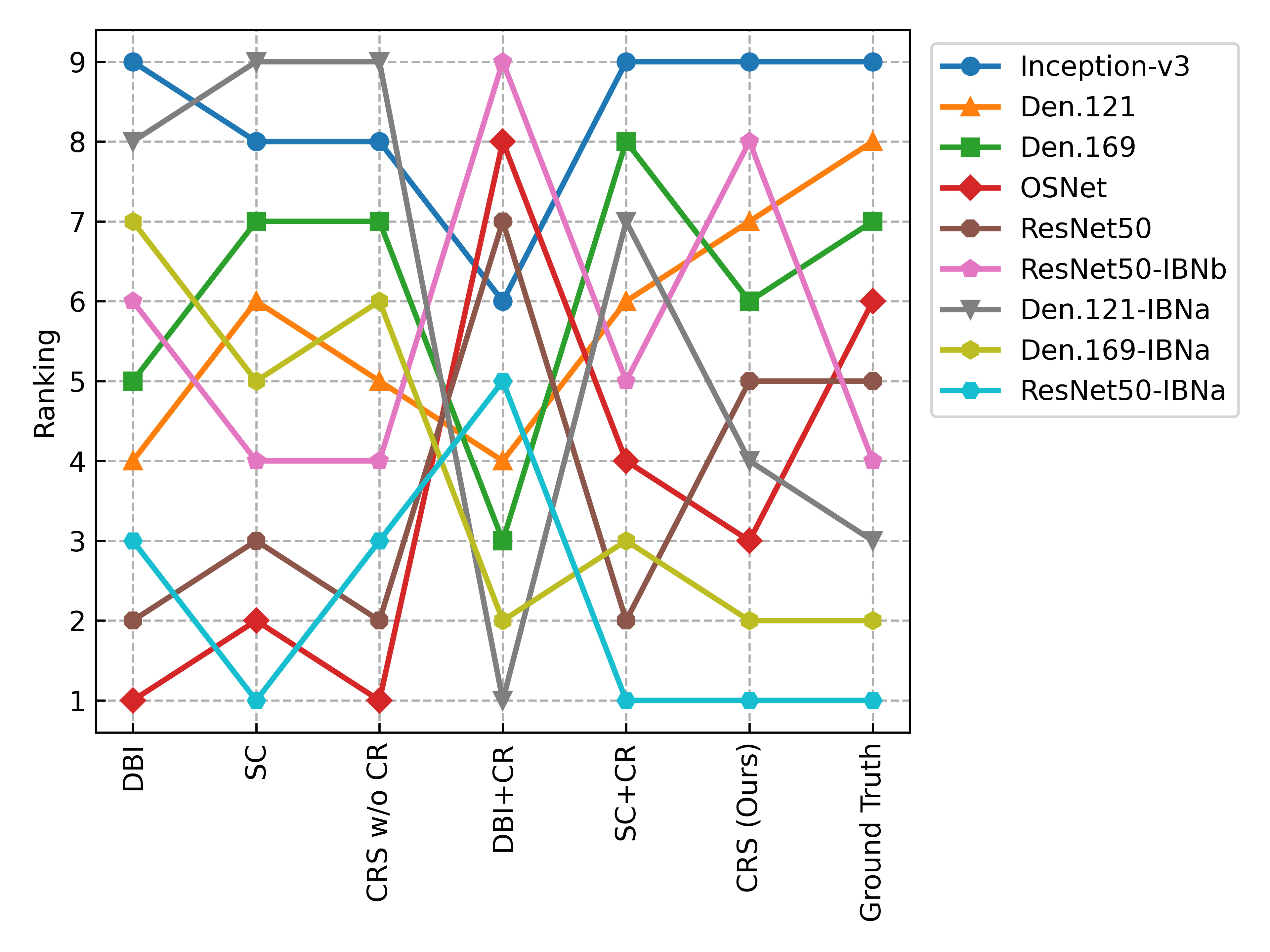}
\caption{Rankings of 9 models under existing clustering measures
and the proposed metric “CRS”. The ground truth ranks models by the mAP after unsupervised learning.}
\label{fig:scatters_rank}
\end{figure}

\begin{figure*}[t]
\centering
\includegraphics[width=1.0\linewidth]{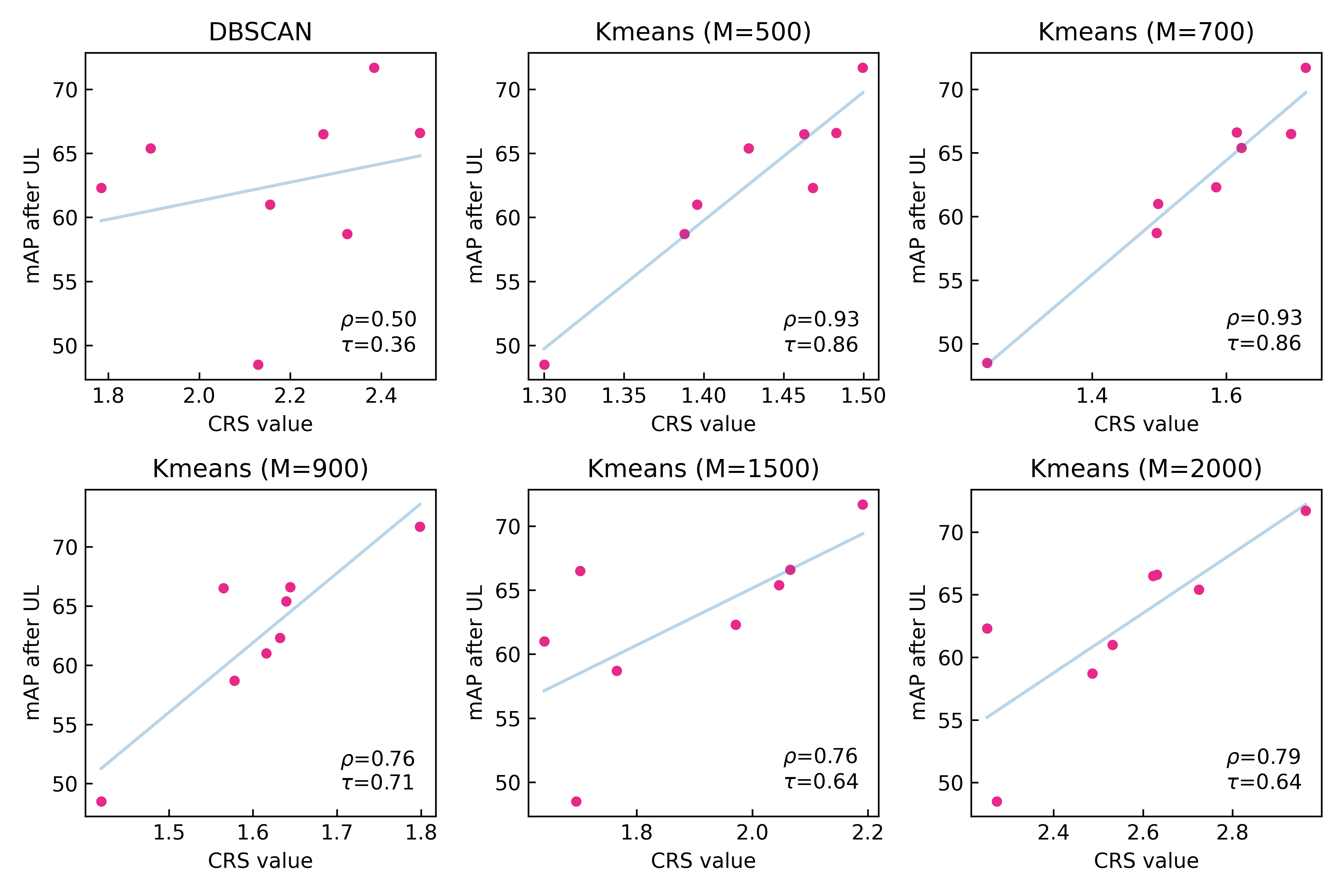}
\caption{Comparison of CRS with different cluster settings on the correlation between the measures and re-ID performance (mAP after unsupervised learning) over different models on Market-1501 dataset. }
\label{fig:scatters_cluster}
\end{figure*}

\begin{table*}[t]
\caption{Comparison of re-ID performance using different cluster numbers $M$ for kmeans clustering in CRS.}
\begin{center}
\small
\begin{tabular}{l|cccc|cccc}
    \hline\hline
    \multirow{2}{*}{Methods} & \multicolumn{4}{c|}{Market-1501} & \multicolumn{4}{c}{DukeMTMC-reID} \\
    \cline{2-9}
    & mAP & R-1 & R-5 & R-10 & mAP & R-1 & R-5 & R-10 \\
    \hline
    PEG (M=500)  & 84.3 &93.7 &97.8 &98.5 &71.9 &83.8 &91.2& 93.5 \\
    PEG (M=700)  & 85.0& 94.1 & 97.8&98.8  & 73.3 &84.8&92.0&94.1 \\
    PEG (M=900)      &  83.8& 93.1 & 97.4& 98.4&72.3&83.7&91.3&93.8 \\
    PEG (M=1500)      &  83.6& 93.1 &97.2 &98.5 &71.5&84.1&91.6&93.6 \\
    PEG (M=2000)      & 83.5& 93.7 &97.8 &98.6 &71.3&83.6&91.2&93.4\\
    \hline\hline
\end{tabular}
\end{center}
\label{table:kmeans}
\end{table*}

\begin{table}[t]
\caption{Comparison with different architecture for the reference model in CRS. All models are trained for 500 iterations during the evaluation of CRS. Models easy to converge such as OSNet and ResNet-50 show better measurement.}
\begin{center}
\begin{tabular}{l|cccc}
    \hline\hline
    Reference Model 
    & $\rho$ & $\tau$ & Param. & Time/iter. \\
    \hline\hline
    DenseNet-121 &0.74&0.57& 6.95M &0.99s\\
    ResNet-50   &0.86&0.71& 23.51M &1.11s\\
    ResNet-101  &0.40&0.36& 42.50M &1.87s\\
    OSNet   &0.93&0.86& 1.91M & 0.98s    \\
    \hline\hline
\end{tabular}
\end{center}
\label{table:crsrm}
\end{table}

\begin{figure*}[h]
\centering
\includegraphics[width=1.0\linewidth]{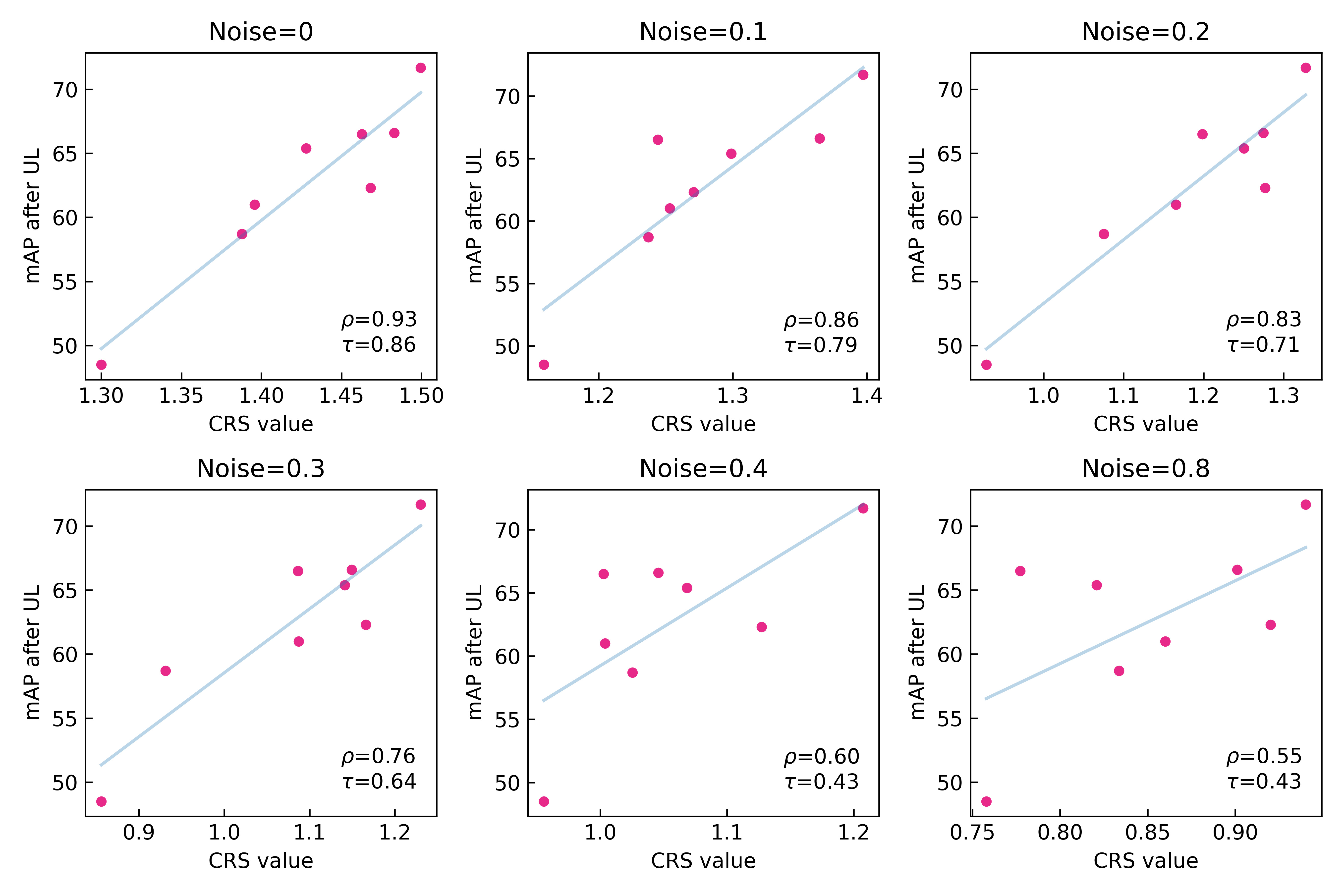}
\caption{Comparison of CRS at different noisy levels of pseudo-labels on the correlation between the measures and re-ID performance (mAP after unsupervised learning) over different models on Market-1501 dataset.}
\label{fig:scatters_noise}
\end{figure*}

\begin{table*}[h]
\caption{Comparison of re-ID performance of PEG using different light-weight networks as reference models in CRS.}
\begin{center}
\begin{tabular}{l|cccc|cccc}
    \hline\hline
    \multirow{2}{*}{Reference Models} & \multicolumn{4}{c|}{Market-1501} & \multicolumn{4}{c}{DukeMTMC-reID} \\
    \cline{2-9}
    & mAP & R-1 & R-5 & R-10 & mAP & R-1 & R-5 & R-10 \\
    \hline
    OSNet  & 84.3 &93.7 &97.8 &98.5 &71.9 &83.8 &91.2& 93.5 \\
    MobileNet  & 83.9&93.1&	97.8&98.6 & 72.0 &83.9&91.2&93.3\\
    ResNet18            &  84.8&93.8&97.7&98.7&72.1&83.4&91.8&93.8\\
    \hline\hline
\end{tabular}
\end{center}
\label{table:refmodel}
\end{table*}

\subsubsection{Evaluation of Cross-Reference Scatter}
\label{sec:expcrs}
In this section, we first validate the basic motivation of the cross-reference scatter, the phenomenon that more accurate labels lead to larger intra-cluster cohesion and inter-cluster separation in the trained feature space. We use inter-/intra-cluster scatters (ICS) to measure the separation as well as the cohesion over the feature space of models. As shown in Fig. \ref{fig:scatters}, a larger ICS means larger inter-cluster separation and intra-cluster cohesion.  
We evaluate the ICS of models trained by labels with different accuracy. Specifically, the label accuracy is controlled by replacing a part of the ground truth with randomly incorrect labels. The results shown in Fig. \ref{fig:lacc} indicate a positive correlation between the ICS and the label accuracy, which confirms the basic hypothesis of CRS. 

We also evaluate the cross-reference scatter with different metrics for clustering algorithm to compare the performance of re-ID models without ground truth. All comparison models with different architectures were first pre-trained in DukeMTMC-reID dataset and then evaluated in Market-1501 by the metrics. To demonstrate whether a metric can show the relative performance between models, we evaluate the correlation between the metric scores and the re-ID performance for all metrics, as shown in Fig. \ref{fig:scatters_arch}. Since CRS measures models at the start of every generation in PEG but aims to select the models that perform better after training, the metric scores were calculated before training and the re-ID performance is evaluated with mAP after the model has been unsupervised trained on the unlabeled data from Market1501, which represents more latent performance of models. We first compare ICS with two metrics for clustering algorithm including Davies-Bouldin Index (DBI) and Silhouette Coefficient (SC), as shown in the first line of Fig. \ref{fig:scatters_arch}. All three metrics are calculated directly on the feature space of the evaluated models by performing k-means clustering. For each metric, we used Spearman’s Rank Correlation ($\rho$) \citep{spearman1961proof} and Kendall’s Rank Correlation ($\tau$) \citep{kendall1938new} to measure the correlation between the metric scores and the re-ID performance. However, we clearly see the poor correlation of the metrics with the re-ID performance according to the small $\rho$ and $\tau$, indicating that the distribution of features before training can not show the real performance of models. 
Then we evaluate the three metrics using our proposed cross-reference (CR) evaluation where metrics are calculated on the feature distribution of a reference model trained by predicted labels. As illustrated in the second line of Fig. \ref{fig:scatters_arch}, correlations are consistently improved by CR, which validates its effectiveness. Importantly, our CRS (ICS+CR) performs the highest correlation with $\rho=0.93$ and $\tau=0.86$ among all six compared metrics. 
Besides, we also present the rankings of models
under different metrics in Fig. \ref{fig:scatters_rank}. Compared with the ground truth ranking result in the last column, CRS achieves a similar ranking of models while other metrics fail to rank them well. 
The superior performance of CRS can be attributed to two reasons. One reason is the cross-reference evaluation that measures the accuracy of predicted labels can better reflect models' performance, and another reason is the ICS which better measures the convergence degree of the reference model. Specifically, both DBI and SC focus on the distribution of the difficult edge samples of clusters while they ignore the overall distribution and thus cannot measure well the degree of model convergence.

{
We also evaluate CRS with different clustering algorithms such as DBSCAN. 
In our work, DBSCAN is adopted in model learning to generate more accurate pseudo-labels like many recent unsupervised re-ID works. However, it is not applicable for CRS because the fair comparison of CRS among models requires the same cluster number during clustering, while DBSCAN cannot guarantee that. Specifically, CRS defined by the ratio of intra-/inter-cluster variance is relative to the cluster numbers. And the cluster numbers by DBSCAN with different evaluated feature models are likely to be different, making it unfair to compare their CRS for model selection. In our work, we use kmeans with the same cluster number \textit{k} for all evaluated models. To validate its effectiveness, we evaluate the correlation between CRS and model performance using different clustering algorithms, as shown in Fig. \ref{fig:scatters_cluster}. 
Compared with kmeans (M=500), DBSCAN achieves a much lower Spearman's Rank Correlation ($\rho$) \citep{spearman1961proof} and Kendall's Rank Correlation ($\tau$) \citep{kendall1938new} between the metric and mAP values. The results show that CRS with DBSCAN fails to measure the models, and KMeans does it better. Therefore, we use kmeans with the same cluster number for CRS.

Furthermore, we study the number of clusters $M$ for k-means in CRS, which is hard to fix in the real world. We first compare the correlation between CRS and re-ID performance with different values of $M$, as shown in Fig. \ref{fig:scatters_cluster}. CRS shows stronger correlations when $M$ is set to 500 or 700, which is close to the number of person IDs in the datasets. The good performance of this cluster number is consistent with other kmeans based methods like MMT \citep{ge2020mutual}. When $M$ is larger, the correlation will be weaker. But the CRS still basically reflects the performance of re-ID models, indicating it is robust to the cluster number.
On the other hand, we also evaluate the performance of our full method with different $M$ on both Market1501 and DukeMTMC-reID datasets. As shown in Table \ref{table:kmeans}, the re-ID performance is generally consistent with the correlations of CRS.
PEG performs best when $M$ is set to 700, where CRS also achieves the highest $\rho$ and $\tau$, making selection able to select better models.
}

{

To validate CRS for very weak evaluated models which predict mostly wrong pseudo labels, we estimate CRS at different noise levels. Although its predicted labels are partially wrong for each evaluated model, we add extra noises by disrupting the label order of a particular portion of samples. As shown in Fig. \ref{fig:scatters_noise}, CRS maintains a stronger correlation between its values and model performance with the increase of the noise ratio, indicating its robustness for the wrong labels. When the noise level is too high such as 0.8, the correlation visibly deteriorated. However, higher CRS can still roughly reflect the better models. 
}

In addition, we evaluate CRS with different architectures of the reference model. Four models are compared with fewer parameters to more parameters including OSNet, DenseNet-121, ResNet-50, and ResNet-101. For fair, all reference models are trained for 500 iterations during the evaluation of CRS. As shown in Table \ref{table:crsrm}, we observe that models easy to converge such as OSNet and ResNet-50 show better measurement for higher correlation $\rho$ and $\tau$, while models hard to converge, like DenseNet-121 and ResNet-101, don't perform well. Specifically, DenseNet using a dynamic architecture and ResNet-101 have deep layers and amounts of parameters, therefore they both require much more time to train. Since only a few training iterations are performed in CRS, the two architectures cannot show a sufficiently differentiable difference in feature distribution when evaluating different models.
{
Moreover, we evaluate PEG for CRS with other light-weight networks as the reference model, including MobileNet and ResNet18. Different from OSNet specially designed for re-ID, the other two architectures are designed for general purpose.
Table \ref{table:refmodel} shows the re-ID performance of PEG with different reference models for CRS. Our method achieves comparable re-ID performance consistently on Market-1501 and DukeMTMC-reID datasets. The results indicate that our CRS metric is model-general for reference models, which is not limited to certain architectures. 
}
In this work, we adopt the OSNet as the reference model of CRS in all other experiments for less time-consuming and more accurate measurement.

\begin{table}[t]
\caption{Comparison with different selection strategies for initial network architectures. The results are tested after once selection and mutual learning without mutation on Market-1501 dataset. }
\begin{center}
\begin{tabular}{l|cccc}
    \hline\hline
    Selection strategies
    & mAP & R-1 & R-5 & R-10  \\
    \hline\hline
    Deepest   & 77.9&89.9&96.1&97.1   \\
    Most heavyweight   & 76.7&89.3&96.1&97.5\\
    \hline  
    Cooperative game (Ours)   &  79.2 & 90.9  & 96.3  & 97.5   \\
    \hline\hline
\end{tabular}
\end{center}
\label{table:expertcommittee}
\end{table}

\begin{table}[t]
\caption{Comparison between individual selection and group selection in PEG. Individual selection selects networks with better individual performance (CRS) while group selection selects the network combination with better overall performance (CRS).}
\begin{center}
\begin{tabular}{c|cccc}
    \hline\hline
    Selection strategies
    & mAP & R-1 & R-5 & R-10  \\
    \hline\hline
    Individual selection   &82.3&92.7&97.1&98.2   \\
    Group selection   &83.6 & 93.3&97.3&98.3   \\ 
    \hline\hline
\end{tabular}
\end{center}
\label{table:ablationgroup}
\end{table}

\begin{figure}[t]
\centering
\includegraphics[width=1.0\linewidth]{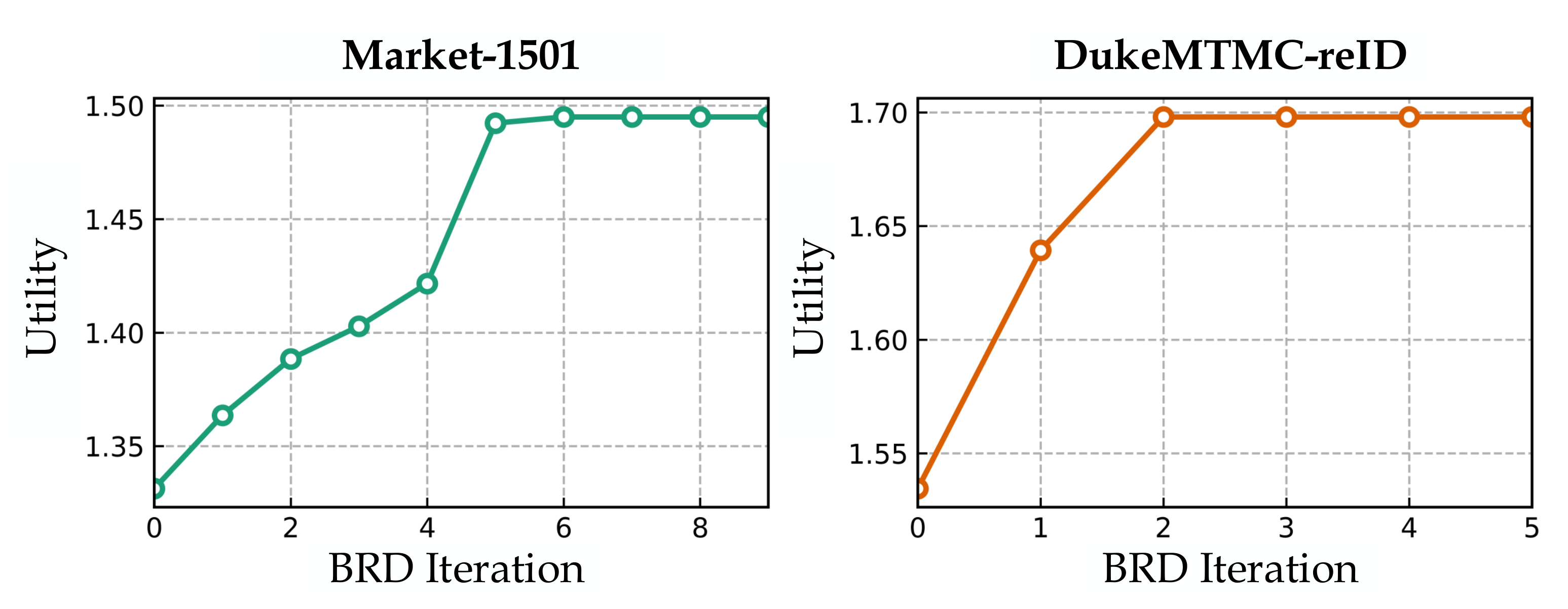}
\caption{Illustration of curves of utility outcome (calculated by CRS) over the best response dynamics (BRD) iterations in the first selection phase during population-based training. The utility outcome increases strictly and eventually halts at a Nash equilibrium.}
\label{fig:brdcurves}
\end{figure}

\begin{figure}[t]
\centering
\includegraphics[width=0.8\linewidth]{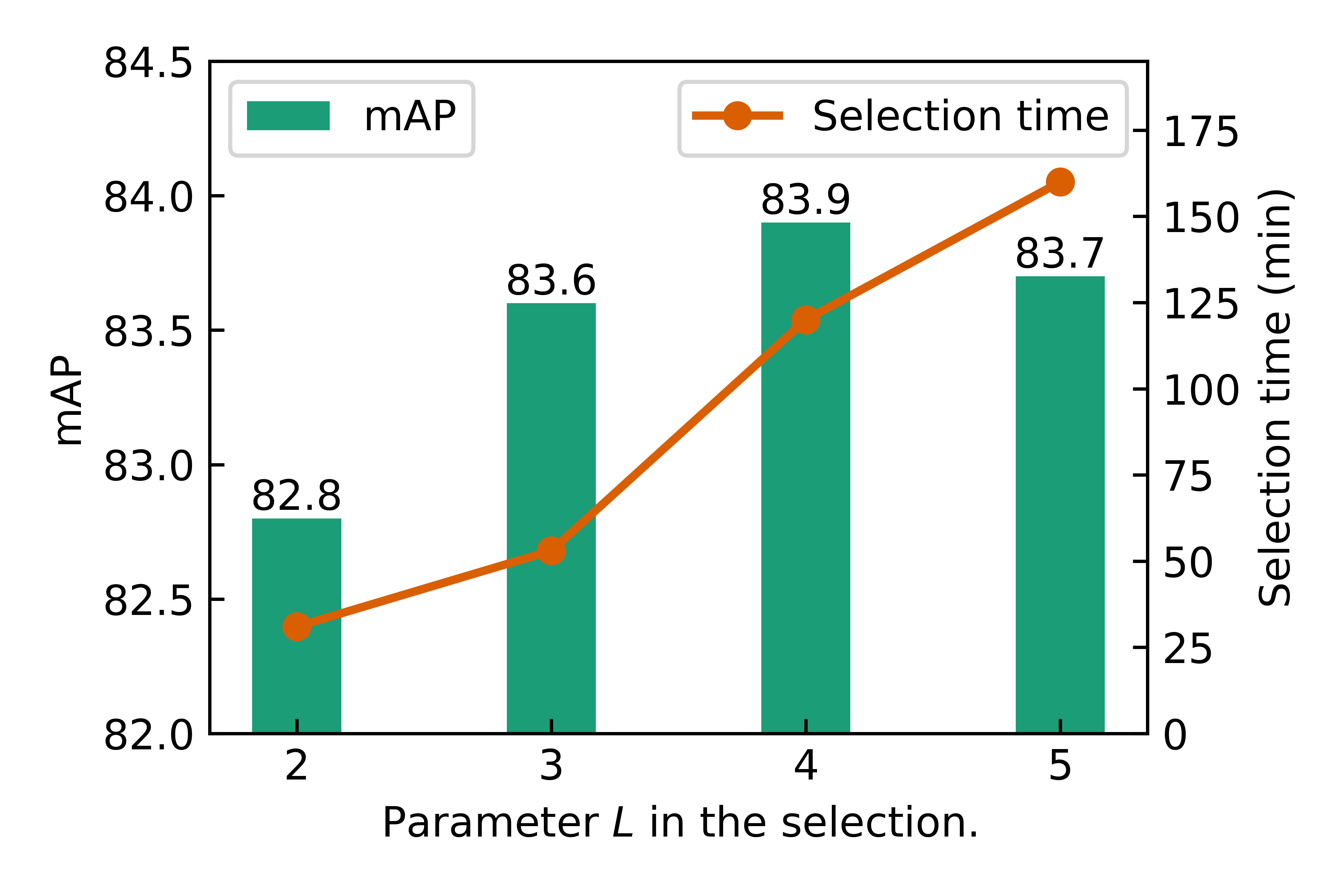}
\caption{Comparison with different agent numbers $L$ in the cooperative selection game. A larger $L$ leads to better performance but higher time consumption. The mutation factor $r$ is set to 0.2 for stability.}
\label{fig:ablation_l}
\end{figure}

\begin{figure}[t]
\centering
\includegraphics[width=0.8\linewidth]{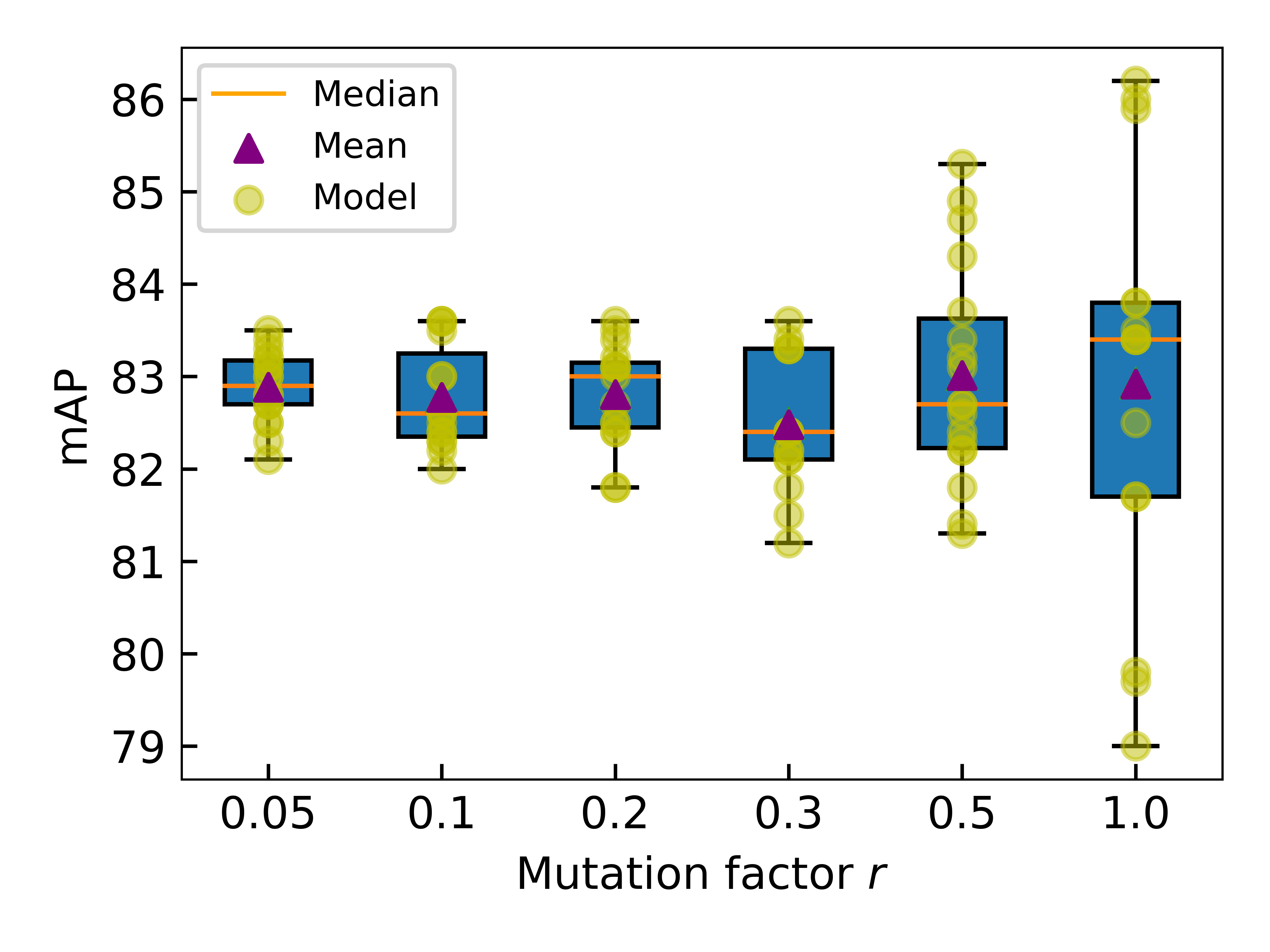}
\caption{Performance of networks in populations over different mutation factor $r$. Each box represents a population and points denote models. A larger $r$ leads to better performance but larger variance of networks within populations.}
\label{fig:rbox}
\end{figure}

\subsubsection{Analysis of Selection}

\textbf{Comparison with different model selection strategies.}
For the method of selection of networks in PEG, we first compare our cooperative gaming (using the best-response dynamics according to CRS) with different selection strategies of network architectures, for example, using some of the deepest or weight-heaviest networks since deeper or weight-heavier networks generally achieve better performance. Considering that these strategies can not select networks from ones with the same architecture, in this experiment, we perform the selection from networks with different architectures only once and then train them by mutual learning before testing. The experiment results shown in Table \ref{table:expertcommittee} indicate that our approach selects better models which achieve higher performance through mutual learning. The better selection can be attributed to that the CRS approximately measures the discriminative capability of models by efficiently using the unlabeled data. 

Moreover, we compare our group selection with the individual selection in PEG. For individual selection, we evaluated the CRS of every single network and accordingly preserved the best $L$ networks. While for group selection, we use the cooperative game to find and preserve the group of $L$ networks with the highest overall CRS. Through the iterative evolutionary game, the group selection performs better, as shown in Table \ref{table:ablationgroup}. The superior performance indicates that networks preserved by the group selection are more complementary and it helps to achieve a better population in later evolutionary training.

\textbf{Convergence analysis of cooperative selection gaming}.
We now discuss the convergence of the cooperative gaming of selection. Note that in every iteration of best response dynamics in Eq. \ref{eq:ops}, the outcome of the utility function strictly increases. Thus, no cycles are possible. Since the game is finite by assumption, it eventually ends, necessarily at a Nash equilibrium. The convergence of the cooperative game is illustrated in Fig. \ref{fig:brdcurves}, where each game eventually halts at a Nash equilibrium.

\subsubsection{Parameter Analysis}

\textbf{Analysis of agent number $L$ in the selection.} The agent number $L$ in the cooperative game of selection determines the maximal size of the selected subset of networks. Here we evaluate the performance of our method and computational cost of the selection over different values of $L$, as shown in Fig. \ref{fig:ablation_l}. Usually, a smaller $L$ will lead to a lack of diversity of the population since only a small number of networks can be preserved during selection. However, $L$ should not be very large because it will waste much more computational resources for solving the best-response dynamics. On the other hand, a larger $L$ also means the larger size of the population in the next generation, which will cost more time for mutual learning among the networks. Taken together, a $L$ of 3 is proper in our experiments, which achieves good performance without consuming too many computational resources.  

\textbf{Analysis of mutation factor $r$.} The mutation factor $r$ in Sec. \ref{sec:mutation} will affect the diversity of populations and so the evolutionary training processes. We studied this parameter by setting it to different values and checking the mAP performance of all networks in the populations. 
{Fig. \ref{fig:rbox} shows experimental results on Market-1501, where each circle point denote a single model.
}
{
Using a larger $r$ usually leads to a higher diversity within populations, which further leads to a higher possibility of achieving better performance. Specifically, a larger r results in a higher upper bound (maximized performance) and a similar average value. 
Notably, the average values do not represent the final performance. 
Although PEG aims to train a population of diverse networks, only one network is selected automatically according to Cross Reference Scatter for inference at the end of training, which is probably to be the better one. Therefore, the final performance of our method doesn't depend on the average values of the population but depends on the performance of the selected model. 
On the other hand, a population with a higher upper bound is more likely to select a better model. For example, when r is set to 0.05 the best model in the population achieves 83.6\% mAP, but when r is set to 0.5, there are 1/4 models in the population that achieve mAP higher than 83.6\%, which may be selected for superior performance. 
{
Importantly, the final model is selected according to Cross Reference Scatter which is to estimate model performance by unlabeled training data. Experimental results in Sec. \ref{sec:expcrs} demonstrated that better models are likely to have higher CRS values to be selected. And when $r=0.5$ it provides more better models as candidates for the final selection.
}
However, a larger $r$ will also bring larger variance and instability of network performance within populations because it may reproduce very weak networks that drag down the overall discrimination capability of the whole population by mutual learning.
Given all of that, we set r to 0.5 for both performance and stability.

}
 
\begin{figure}[t]
\centering
\begin{minipage}[b]{1\linewidth}
\centering
\small
\begin{tabular}{c|ccccc}
    \hline
    Generation & 1 & 2 &3 &4 &5  \\
    \hline
    mAP & 66.0 & 83.8 & 84.9 & 85.3 & 84.4 \\
    R-1 & 81.8 & 93.0 & 93.5 & 94.2 &  93.1 \\
    \hline
\end{tabular}
\end{minipage} 
\centering
\begin{minipage}[b]{1\linewidth}
    \centering
    \includegraphics[width=0.8\linewidth]{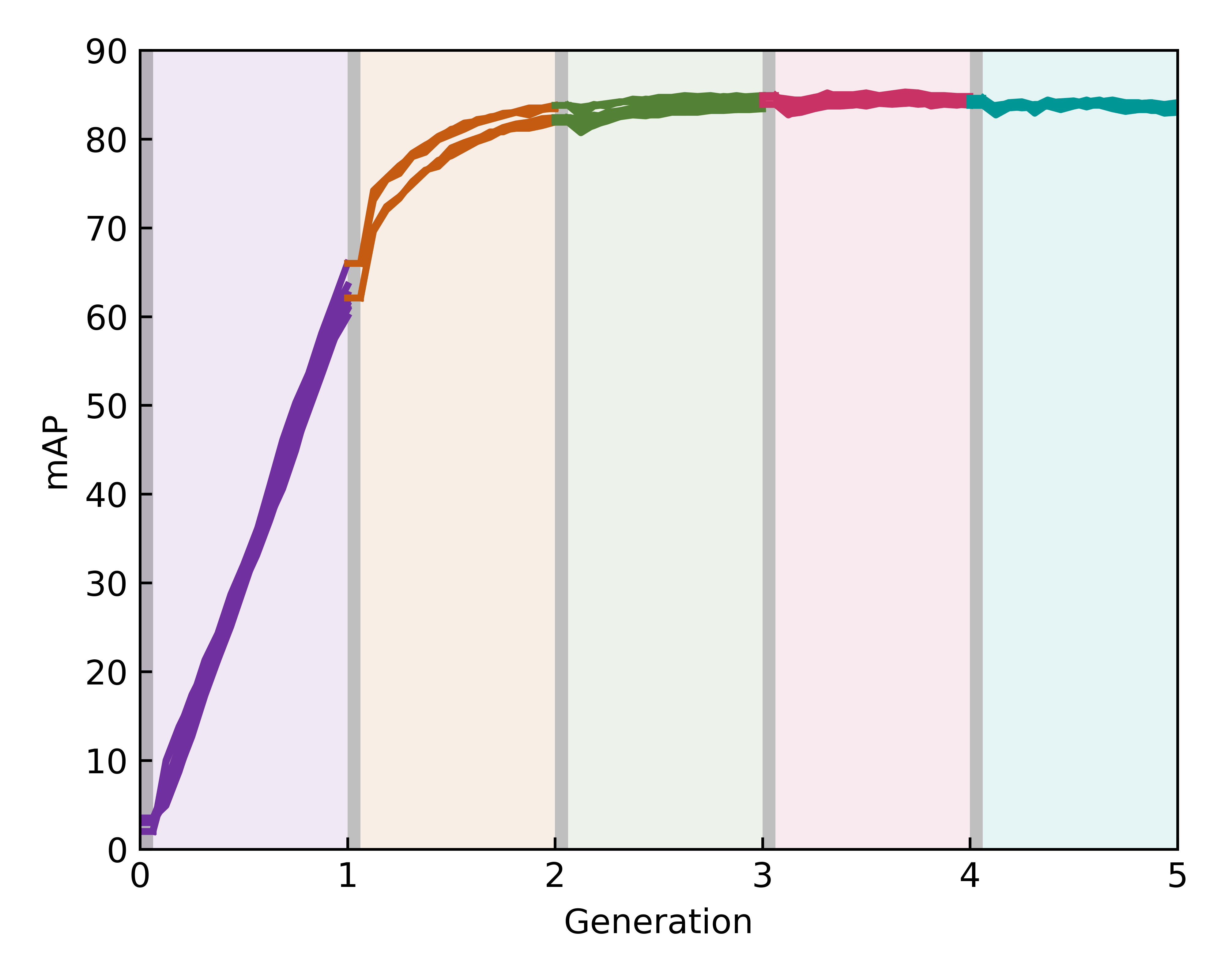}
\end{minipage} 
\caption{Illustration of the model performance in every generation for 5 generation evolution.}
\label{fig:peg_gen}
\end{figure}

{
\textbf{Analysis of the number of generations.}
To analyze the number of generations, we provide the model performance in every generation for 5 generation evolution, as shown in Fig. \ref{fig:peg_gen}. The performance of models is boosted rapidly in the first and second generation, and the boost slows down gradually as the generation increases. After three generations of evolution, the performance is nearly convergent, and models achieve stable results. 
}

\begin{figure}[t]
\centering
\includegraphics[width=1.0\linewidth]{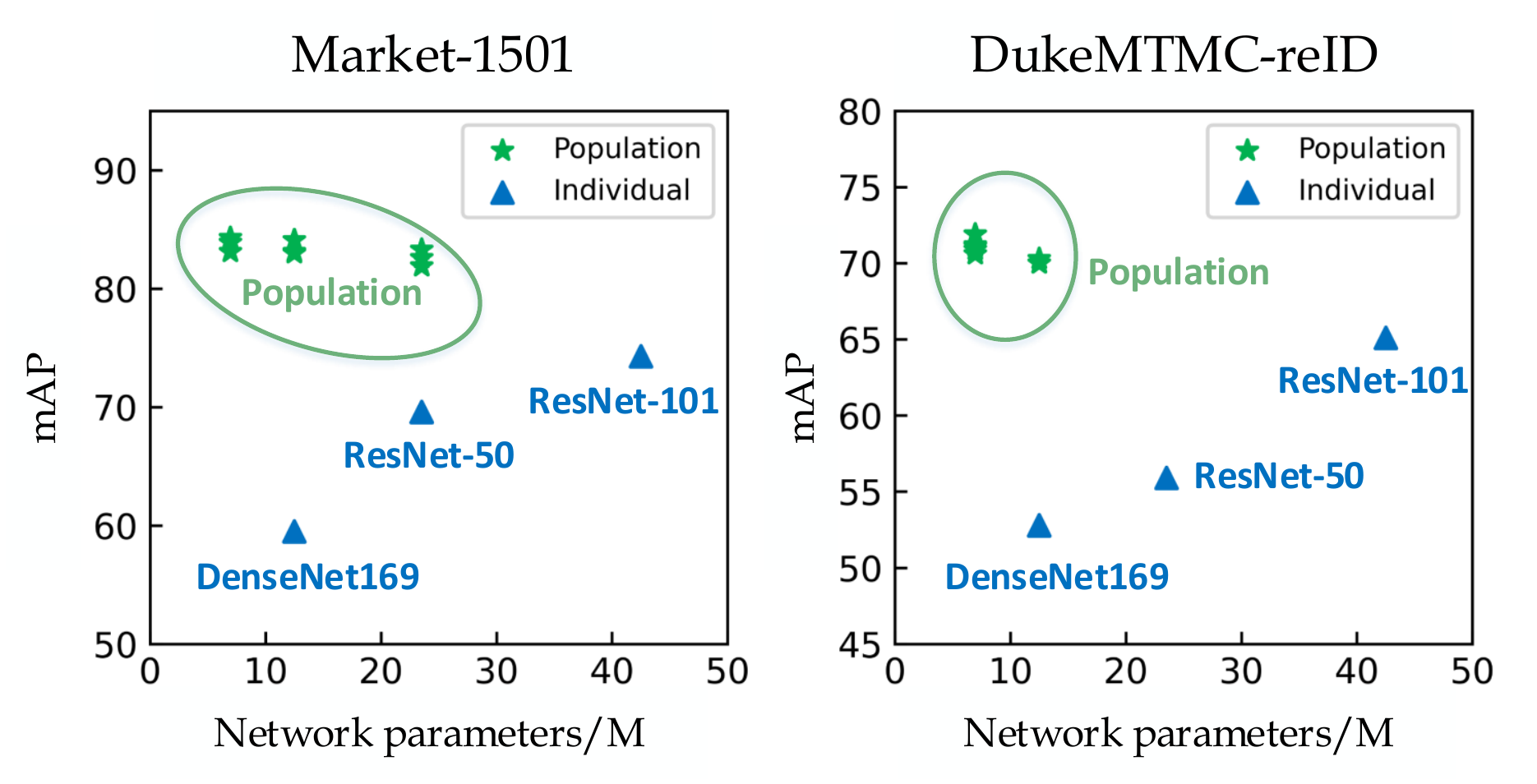}
\caption{Comparison between the lightweight networks within the population and heavyweight networks trained individually on two datasets.}
\label{fig:heavyweight}
\end{figure}

\begin{table*}[t]
\caption{Comparison of computational cost between single models and PEG with different population sizes. PEG (small) selects only two models during selection, and every model reproduces to 2 times. And PEG (large) follows the original settings according to the implementation details. }
\begin{center}
\footnotesize
\begin{tabular}{l|cc|ccc|cc}
    \hline\hline
    \multirow{2}{*}{Methods} & \multicolumn{2}{c|}{Performance} & \multicolumn{3}{c|}{Training cost} & \multicolumn{2}{c}{Testing cost} \\
    \cline{2-8}
    & mAP & R-1 & Param. /M & Complexity /GMac & Time /h & Param. /M & Complexity /GMac \\
    \hline\hline
    IBN-DenseNet169   & 57.4  & 75.2  &  12.49 & 2.23  & 3.5 & 12.49 & 2.23    \\
    IBN-ResNet50   & 69.6 & 84.9 &  23.51 & 4.08  & 3.9 &  23.51 & 4.08   \\
    IBN-ResNext101   & 72.2 & 87.4 & 42.13  & 6.54  & 4.8 & 42.13  & 6.54   \\
    ResNet200   & 73.1 & 88.7 & 62.65  & 10.02  & 8.9 & 62.65  & 10.02   \\
    ResNetrs420   & 73.1 & 86.5 & 189.84  & 20.62  & 19.3 & 189.84  & 20.62   \\
    \hline
    PEG (small)   & 84.1 & 93.0   &  72.00 & 12.62  & 10.7 &  12.49 & 2.23     \\
    PEG (large)   & 84.3 & 93.7 & 128.85 & 24.57 &21.1 & 12.49 & 2.23     \\
    \hline\hline
\end{tabular}
\end{center}
\label{table:cost}
\end{table*}

\begin{figure*}[t]
\begin{minipage}[b]{1\linewidth}
\centering
\begin{tabular}{c|c|c|c}
    \hline
    Selection time/h & Clustering time/h & Model learning time/h & Total time/h \\
    \hline
    3.39 & 1.27 & 17.7 & 21.14 \\ 
    \hline
\end{tabular}
\end{minipage} 
\centering
\begin{minipage}[b]{1\linewidth}
    \centering
    \includegraphics[width=0.7\linewidth]{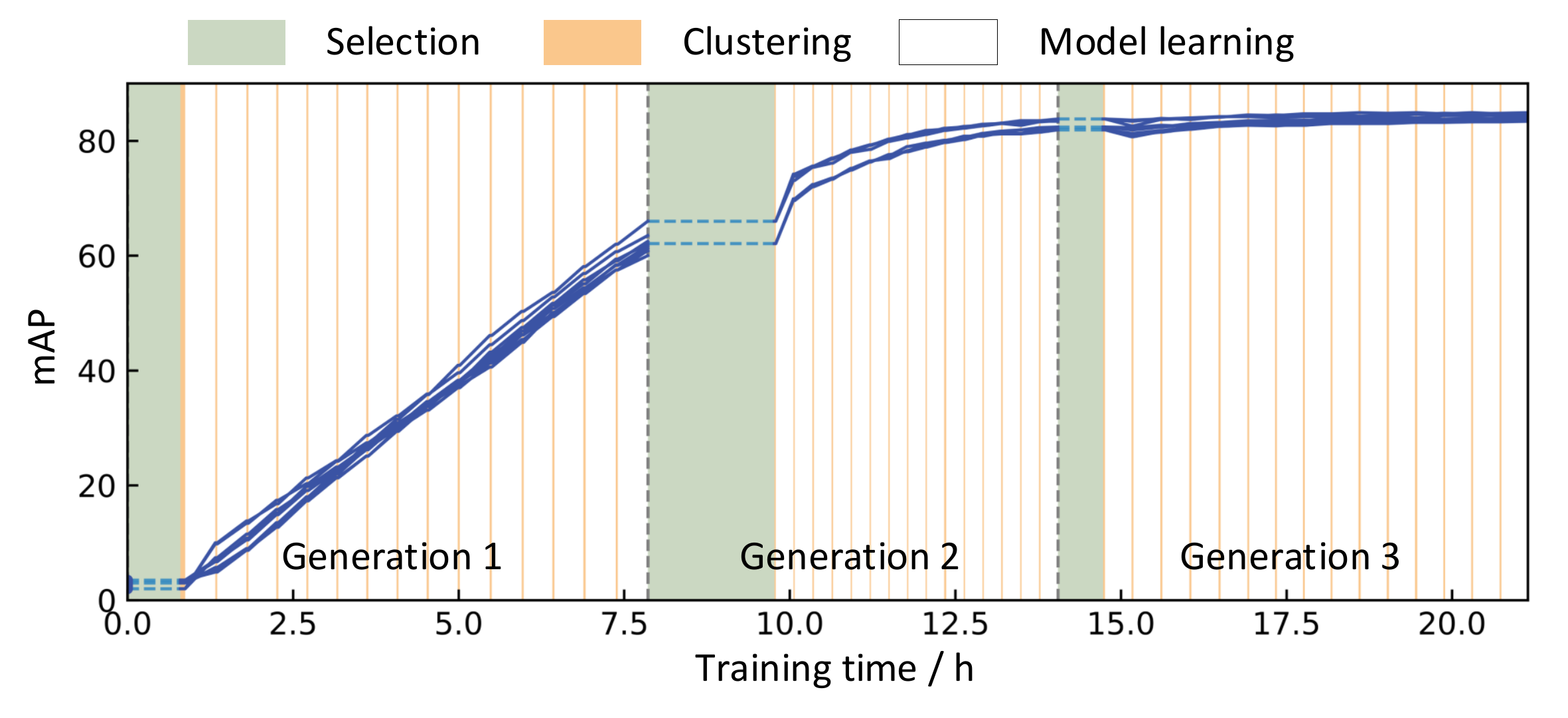}
\end{minipage} 
\caption{Illustration of time consumption of every procedure in our approach. Time of selection, clustering and model learning are represented by green, yellow and white, respectively. Blue curves represent the performance of every model in the population.}
\label{fig:time_slots3}
\end{figure*}

\subsubsection{Multiple models vs. heavyweight models}

Heavyweight networks are more likely to learn discriminative representations than lightweight models since they have deeper architectures and more parameters. However, models with heavyweights require more time and computational resources during both the training and inference stages, making them infeasible in practice. Our experiments show that through PEG, lightweight models can surpass heavyweight models with IBN that are individually trained under unsupervised conditions. Take the market-1501 dataset as an example. After the evolutionary game of the population of lightweight networks, all member networks of the population achieve better performance than heavyweight networks, such as ResNet-101, as shown in Fig. \ref{fig:heavyweight}. Specifically, one of the networks achieves a much higher mAP of 84.3\% than ResNet-101 with only 1/3 of the parameters. The superior results can be attributed to two aspects. One is the mutation and selection that sufficiently explore and exploit the population. Mutation makes networks learn diverse knowledge, and selection maintains the optimal model groups and abandons the others. The models in the selected and preserved groups are complementary, so they produce more accurate and robust pseudo-labels for the next training phase and learn more discriminative features. The second reason is the mutual learning performed among all networks in the whole population. Since the models preserved by mutation and selection are diverse and complementary, each contains only a small part of the knowledge of the whole population. Through mutual learning, the knowledge of the population is assembled into each network by distillation, which equips the models with more discriminative capability. The PEG method explores the potential of lightweight networks and searches for the approximate global optimal solution and thus outperforms the heavyweight models.

{

\subsection{Computational Cost}

To evaluate the efficiency and effectiveness of the computational cost, we evaluate a series of large single models for reference as shown in Table \ref{table:cost}. Experiments are conducted on four V100 GPUs. From the perspective of training, our method requires comparable computational cost with the large single models (such as ResNetrs420) while achieving significant performance improvement. And from the perspective of testing, our method requires as less computational cost as the small single models (such as IBN-DenseNet169) and surpasses them by large margins. More detailed descriptions are listed below.

\textbf{(1) The improvement by increasing parameters is limited for single models on re-ID performance, and our PEG largely surpasses the best single model with comparable computations, demonstrating the cost meaningful.} Specifically, we evaluate five architectures from lightweight to heavyweight including IBN-DenseNet169, IBN-ResNet50, IBN-ResNext101, ResNet200 and ResNetrs420. As parameters increase, single models usually achieve better performance, whereas they require more computational complexity and time for training. However, the improvement is limited when the parameters are very large, i.e., ResNetrs420 cannot surpass ResNet200 even though more than two times of parameters and training time are used. 
Compared with single models, PEG improves the accuracy by large margins. Although population-based training demands more cost, the cost is worth and affordable. 
Importantly, PEG provides further improvement that cannot be achieved by simply increasing model parameters. 

To reduce the cost, we provide two implementation versions of PEG: small and large, with different sizes of the population. Specifically, PEG (small) maintains a lightweight population for efficient training, which selects only two models during selection and every model reproduces to 2 times. And PEG (large) follows the original settings according to the implementation details. Significantly, the provided PEG (small) achieves comparable performance with PEG (large) and requires only half the training cost. It also outperforms the best single architecture ResNet200 for more than 10\% of mAP while costing comparable computing resources, which is more affordable and efficient than the large version.
We suggest the version of PEG (small) for application in resource-limited environments. 

\textbf{(2) PEG requires less computational cost for testing, making it more applicable and valuable in practice.} As is shown in Table \ref{table:cost}, the computational complexity during testing of PEG is largely less than the large single models, even nearly 1/5 of the best one, ResNet200. It only requires as less computational cost as small single models such as IBN-DenseNet169 while surpassing its performance by large margins. It is because only one network in the population is selected in the end for evaluation. Since the training procedure is only conducted once, while the test will be continuously repeated in the actual re-ID system, PEG with less testing cost is applicable and valuable in practice. 
}

{
To further analyze the training time of every procedure in our approach, we illustrated the training process over time in Fig. \ref{fig:time_slots3}. Among the total training time, model learning accounts for the largest proportion. Model learning is performed by data loading, feedforward of all networks, backward of losses, and updating of parameters. This part of time is relative to the number and depth of networks. For example, the time of model learning in Generation 2 is shorter than in the other generations because there are only four networks in the population. The time of the selection stage is different in the three generations. On the one hand, it is affected by the number of candidate networks. On the other hand, it is affected by the convergence of the best-response dynamics. Moreover, clustering costs the least time, only for the extraction of features and execution of clustering algorithms.
}

\section{Conclusion}
The paper proposed a population-based evolutionary gaming which trains concurrently a population of networks for unsupervised person re-ID. We demonstrate that the population can evolve and achieve progressive discrimination through iterative selection to preserve adaptive networks, reproduction and mutation to provide more diversity, and mutual learning to assemble knowledge. Moreover, our proposed cross-reference scatter can approximately estimate the performance of networks using unlabeled data and thus is utilized as the utility of cooperative game in the selection phase. 
Our approach not only produces a new state-of-the-art accuracy on multiple benchmarks but also provided a fresh insight for population-based multi-network training. 

\noindent\textbf{Data availability statement.} The datasets generated during and/or analysed during the current study are available from the corresponding author on reasonable request.

\section*{Acknowledgments}
This work is partially supported by grants from the Key-Area Research and Development Program of Guangdong Province under contact No.2019B010153002, and grants from the National Natural Science Foundation of China under contract No. 61825101 and No. 62088102. The computing resources of Pengcheng Cloudbrain are used in this research.

\bibliographystyle{spbasic} 
\bibliography{egbib}

\begin{thebibliography}{92}
\providecommand{\natexlab}[1]{#1}
\providecommand{\url}[1]{{#1}}
\providecommand{\urlprefix}{URL }
\expandafter\ifx\csname urlstyle\endcsname\relax
  \providecommand{\doi}[1]{DOI~\discretionary{}{}{}#1}\else
  \providecommand{\doi}{DOI~\discretionary{}{}{}\begingroup
  \urlstyle{rm}\Url}\fi
\providecommand{\eprint}[2][]{\url{#2}}

\bibitem[{Ali et~al.(2020)Ali, Moriyama, Kalintha, Numao, and
  Fukui}]{ali2020reinforcement}
Ali B, Moriyama K, Kalintha W, Numao M, Fukui KI (2020) Reinforcement learning
  based metric filtering for evolutionary distance metric learning. Intelligent
  Data Analysis 24(6):1345--1364

\bibitem[{Baker and Hubert(1975)}]{baker1975measuring}
Baker FB, Hubert LJ (1975) Measuring the power of hierarchical cluster
  analysis. Journal of the American Statistical Association 70(349):31--38

\bibitem[{Caron et~al.(2018)Caron, Bojanowski, Joulin, and
  Douze}]{caron2018deep}
Caron M, Bojanowski P, Joulin A, Douze M (2018) Deep clustering for
  unsupervised learning of visual features. In: Proceedings of the European
  Conference on Computer Vision (ECCV), pp 132--149

\bibitem[{Chen et~al.(2020)Chen, Lu, Lu, and Zhou}]{chen2020deep}
Chen G, Lu Y, Lu J, Zhou J (2020) Deep credible metric learning for
  unsupervised domain adaptation person re-identification. In: Computer
  Vision--ECCV 2020: 16th European Conference, Glasgow, UK, August 23--28,
  2020, Proceedings, Part VIII 16, Springer, pp 643--659

\bibitem[{Chen et~al.(2021{\natexlab{a}})Chen, Lagadec, and
  Bremond}]{chen2021ice}
Chen H, Lagadec B, Bremond F (2021{\natexlab{a}}) Ice: Inter-instance
  contrastive encoding for unsupervised person re-identification. arXiv
  preprint arXiv:210316364

\bibitem[{Chen et~al.(2021{\natexlab{b}})Chen, Wang, Lagadec, Dantcheva, and
  Bremond}]{chen2021joint}
Chen H, Wang Y, Lagadec B, Dantcheva A, Bremond F (2021{\natexlab{b}}) Joint
  generative and contrastive learning for unsupervised person
  re-identification. In: Proceedings of the IEEE/CVF Conference on Computer
  Vision and Pattern Recognition, pp 2004--2013

\bibitem[{Dai et~al.(2021)Dai, Wang, Zhu, Yuan, and Tan}]{dai2021cluster}
Dai Z, Wang G, Zhu S, Yuan W, Tan P (2021) Cluster contrast for unsupervised
  person re-identification. arXiv preprint arXiv:210311568

\bibitem[{Davies and Bouldin(1979)}]{4766909}
Davies DL, Bouldin DW (1979) A cluster separation measure. IEEE Transactions on
  Pattern Analysis and Machine Intelligence PAMI-1(2):224--227,
  \doi{10.1109/TPAMI.1979.4766909}

\bibitem[{Deng et~al.(2009)Deng, Dong, Socher, Li, Li, and
  Li}]{DBLP:conf/cvpr/DengDSLL009}
Deng J, Dong W, Socher R, Li L, Li K, Li F (2009) Imagenet: {A} large-scale
  hierarchical image database. In: IEEE CVPR

\bibitem[{Deng et~al.(2018)Deng, Zheng, Ye, Kang, Yang, and
  Jiao}]{Deng_2018_CVPR}
Deng W, Zheng L, Ye Q, Kang G, Yang Y, Jiao J (2018) Image-image domain
  adaptation with preserved self-similarity and domain-dissimilarity for person
  re-identification. In: IEEE CVPR

\bibitem[{Dietterich(2000)}]{dietterich2000ensemble}
Dietterich TG (2000) Ensemble methods in machine learning. In: International
  workshop on multiple classifier systems, Springer, pp 1--15

\bibitem[{Dunn(1973)}]{dunn1973fuzzy}
Dunn JC (1973) A fuzzy relative of the isodata process and its use in detecting
  compact well-separated clusters

\bibitem[{Ester et~al.(1996)Ester, Kriegel, Sander, and
  Xu}]{DBLP:conf/kdd/EsterKSX96}
Ester M, Kriegel H, Sander J, Xu X (1996) A density-based algorithm for
  discovering clusters in large spatial databases with noise. In: KDD, pp
  226--231

\bibitem[{Fan et~al.(2018)Fan, Zheng, Yan, and
  Yang}]{DBLP:journals/tomccap/FanZYY18}
Fan H, Zheng L, Yan C, Yang Y (2018) Unsupervised person re-identification:
  Clustering and fine-tuning. {TOMCCAP} 14(4):83:1--83:18

\bibitem[{Fu et~al.(2021)Fu, Chen, Bao, Yang, Yuan, Zhang, Li, and
  Chen}]{fu2021unsupervised}
Fu D, Chen D, Bao J, Yang H, Yuan L, Zhang L, Li H, Chen D (2021) Unsupervised
  pre-training for person re-identification. In: Proceedings of the IEEE/CVF
  Conference on Computer Vision and Pattern Recognition, pp 14750--14759

\bibitem[{Fu et~al.(2019)Fu, Wei, Wang, Zhou, Shi, and Huang}]{fu2019self}
Fu Y, Wei Y, Wang G, Zhou Y, Shi H, Huang TS (2019) Self-similarity grouping: A
  simple unsupervised cross domain adaptation approach for person
  re-identification. In: Proceedings of the IEEE International Conference on
  Computer Vision (ICCV), pp 6112--6121

\bibitem[{Fukui et~al.(2013)Fukui, Ono, Megano, and
  Numao}]{fukui2013evolutionary}
Fukui Ki, Ono S, Megano T, Numao M (2013) Evolutionary distance metric learning
  approach to semi-supervised clustering with neighbor relations. In: 2013 IEEE
  25th International Conference on Tools with Artificial Intelligence, IEEE, pp
  398--403

\bibitem[{Ge et~al.(2020{\natexlab{a}})Ge, Chen, and Li}]{ge2020mutual}
Ge Y, Chen D, Li H (2020{\natexlab{a}}) Mutual mean-teaching: Pseudo label
  refinery for unsupervised domain adaptation on person re-identification.
  arXiv preprint arXiv:200101526

\bibitem[{Ge et~al.(2020{\natexlab{b}})Ge, Zhu, Chen, Zhao, and
  Li}]{ge2020self}
Ge Y, Zhu F, Chen D, Zhao R, Li H (2020{\natexlab{b}}) Self-paced contrastive
  learning with hybrid memory for domain adaptive object re-id. arXiv preprint
  arXiv:200602713

\bibitem[{Goodfellow et~al.(2014)Goodfellow, Pouget-Abadie, Mirza, Xu,
  Warde-Farley, Ozair, Courville, and Bengio}]{goodfellow2014generative}
Goodfellow I, Pouget-Abadie J, Mirza M, Xu B, Warde-Farley D, Ozair S,
  Courville A, Bengio Y (2014) Generative adversarial nets. In: Advances in
  neural information processing systems, pp 2672--2680

\bibitem[{Halkidi et~al.(2002)Halkidi, Batistakis, and
  Vazirgiannis}]{halkidi2002clustering}
Halkidi M, Batistakis Y, Vazirgiannis M (2002) Clustering validity checking
  methods: Part ii. ACM Sigmod Record 31(3):19--27

\bibitem[{Hansen and Salamon(1990)}]{hansen1990neural}
Hansen LK, Salamon P (1990) Neural network ensembles. IEEE transactions on
  pattern analysis and machine intelligence 12(10):993--1001

\bibitem[{Ho et~al.(2019)Ho, Liang, Chen, Stoica, and
  Abbeel}]{ho2019population}
Ho D, Liang E, Chen X, Stoica I, Abbeel P (2019) Population based augmentation:
  Efficient learning of augmentation policy schedules. In: International
  Conference on Machine Learning, PMLR, pp 2731--2741

\bibitem[{Huang et~al.(2016)Huang, Sun, Liu, Sedra, and
  Weinberger}]{huang2016deep}
Huang G, Sun Y, Liu Z, Sedra D, Weinberger KQ (2016) Deep networks with
  stochastic depth. In: European conference on computer vision (ECCV),
  Springer, pp 646--661

\bibitem[{Huang et~al.(2017{\natexlab{a}})Huang, Li, Pleiss, Liu, Hopcroft, and
  Weinberger}]{huang2017snapshot}
Huang G, Li Y, Pleiss G, Liu Z, Hopcroft JE, Weinberger KQ (2017{\natexlab{a}})
  Snapshot ensembles: Train 1, get m for free. arXiv preprint arXiv:170400109

\bibitem[{Huang et~al.(2017{\natexlab{b}})Huang, Liu, Van Der~Maaten, and
  Weinberger}]{huang2017densely}
Huang G, Liu Z, Van Der~Maaten L, Weinberger KQ (2017{\natexlab{b}}) Densely
  connected convolutional networks. In: Proceedings of the IEEE conference on
  computer vision and pattern recognition (CVPR), pp 4700--4708

\bibitem[{Huang et~al.(2019)Huang, Peng, Jin, Xing, Lang, and
  Feng}]{huang2019domain}
Huang Y, Peng P, Jin Y, Xing J, Lang C, Feng S (2019) Domain adaptive attention
  model for unsupervised cross-domain person re-identification. arXiv preprint
  arXiv:190510529

\bibitem[{Hubert and Levin(1976)}]{hubert1976general}
Hubert LJ, Levin JR (1976) A general statistical framework for assessing
  categorical clustering in free recall. Psychological bulletin 83(6):1072

\bibitem[{Jaderberg et~al.(2017)Jaderberg, Dalibard, Osindero, Czarnecki,
  Donahue, Razavi, Vinyals, Green, Dunning, Simonyan
  et~al.}]{jaderberg2017population}
Jaderberg M, Dalibard V, Osindero S, Czarnecki WM, Donahue J, Razavi A, Vinyals
  O, Green T, Dunning I, Simonyan K, et~al. (2017) Population based training of
  neural networks. arXiv preprint arXiv:171109846

\bibitem[{Jaderberg et~al.(2019)Jaderberg, Czarnecki, Dunning, Marris, Lever,
  Castaneda, Beattie, Rabinowitz, Morcos, Ruderman et~al.}]{jaderberg2019human}
Jaderberg M, Czarnecki WM, Dunning I, Marris L, Lever G, Castaneda AG, Beattie
  C, Rabinowitz NC, Morcos AS, Ruderman A, et~al. (2019) Human-level
  performance in 3d multiplayer games with population-based reinforcement
  learning. Science 364(6443):859--865

\bibitem[{Ji et~al.(2021)Ji, Wang, Zhou, Tang, Zheng, and Hua}]{ji2021meta}
Ji H, Wang L, Zhou S, Tang W, Zheng N, Hua G (2021) Meta pairwise relationship
  distillation for unsupervised person re-identification. In: Proceedings of
  the IEEE/CVF International Conference on Computer Vision, pp 3661--3670

\bibitem[{Jin et~al.(2020)Jin, Lan, Zeng, and Chen}]{jin2020global}
Jin X, Lan C, Zeng W, Chen Z (2020) Global distance-distributions separation
  for unsupervised person re-identification. arXiv preprint arXiv:200600752

\bibitem[{Kalintha et~al.(2019)Kalintha, Ono, Numao, and
  Fukui}]{kalintha2019kernelized}
Kalintha W, Ono S, Numao M, Fukui Ki (2019) Kernelized evolutionary distance
  metric learning for semi-supervised clustering. Intelligent Data Analysis
  23(6):1271--1297

\bibitem[{Kendall(1938)}]{kendall1938new}
Kendall MG (1938) A new measure of rank correlation. Biometrika 30(1/2):81--93

\bibitem[{Kingma and Ba(2014)}]{kingma2014adam}
Kingma DP, Ba J (2014) Adam: A method for stochastic optimization. arXiv
  preprint arXiv:14126980

\bibitem[{Krogh and Vedelsby(1994)}]{krogh1994neural}
Krogh A, Vedelsby J (1994) Neural network ensembles, cross validation, and
  active learning. Advances in neural information processing systems 7:231--238

\bibitem[{Lakshminarayanan et~al.(2017)Lakshminarayanan, Pritzel, and
  Blundell}]{lakshminarayanan2017simple}
Lakshminarayanan B, Pritzel A, Blundell C (2017) Simple and scalable predictive
  uncertainty estimation using deep ensembles. In: Advances in neural
  information processing systems, pp 6402--6413

\bibitem[{Li and Zhang(2020)}]{li2020joint}
Li J, Zhang S (2020) Joint visual and temporal consistency for unsupervised
  domain adaptive person re-identification. In: European Conference on Computer
  Vision, Springer, pp 483--499

\bibitem[{Li et~al.(2018)Li, Zhu, and Gong}]{li2018unsupervised}
Li M, Zhu X, Gong S (2018) Unsupervised person re-identification by deep
  learning tracklet association. In: Proceedings of the European conference on
  computer vision (ECCV), pp 737--753

\bibitem[{Li et~al.(2019{\natexlab{a}})Li, Zhu, and Gong}]{li2019unsupervised}
Li M, Zhu X, Gong S (2019{\natexlab{a}}) Unsupervised tracklet person
  re-identification. IEEE transactions on pattern analysis and machine
  intelligence 42(7):1770--1782

\bibitem[{Li et~al.(2019{\natexlab{b}})Li, Lin, Lin, and Wang}]{li2019cross}
Li YJ, Lin CS, Lin YB, Wang YCF (2019{\natexlab{b}}) Cross-dataset person
  re-identification via unsupervised pose disentanglement and adaptation. In:
  Proceedings of the IEEE International Conference on Computer Vision (ICCV),
  pp 7919--7929

\bibitem[{Liao et~al.(2015)Liao, Hu, Zhu, and Li}]{Liao_2015_CVPR}
Liao S, Hu Y, Zhu X, Li SZ (2015) Person re-identification by local maximal
  occurrence representation and metric learning. In: The IEEE Conference on
  Computer Vision and Pattern Recognition (CVPR)

\bibitem[{Lin et~al.(2019)Lin, Dong, Zheng, Yan, and Yang}]{lin2019bottom}
Lin Y, Dong X, Zheng L, Yan Y, Yang Y (2019) A bottom-up clustering approach to
  unsupervised person re-identification. In: Proceedings of the AAAI Conference
  on Artificial Intelligence, vol~33, pp 8738--8745

\bibitem[{Lin et~al.(2020)Lin, Xie, Wu, Yan, and Tian}]{lin2020unsupervised}
Lin Y, Xie L, Wu Y, Yan C, Tian Q (2020) Unsupervised person re-identification
  via softened similarity learning. In: Proceedings of the IEEE/CVF Conference
  on Computer Vision and Pattern Recognition, pp 3390--3399

\bibitem[{Liu et~al.(2019)Liu, Zha, Chen, Hong, and Wang}]{Liu_2019_CVPR}
Liu J, Zha ZJ, Chen D, Hong R, Wang M (2019) Adaptive transfer network for
  cross-domain person re-identification. In: IEEE CVPR

\bibitem[{Maulik and Bandyopadhyay(2002)}]{maulik2002performance}
Maulik U, Bandyopadhyay S (2002) Performance evaluation of some clustering
  algorithms and validity indices. IEEE Transactions on pattern analysis and
  machine intelligence 24(12):1650--1654

\bibitem[{Pan et~al.(2018)Pan, Luo, Shi, and Tang}]{pan2018two}
Pan X, Luo P, Shi J, Tang X (2018) Two at once: Enhancing learning and
  generalization capacities via ibn-net. In: Proceedings of the European
  Conference on Computer Vision (ECCV), pp 464--479

\bibitem[{Peng et~al.(2016)Peng, Xiang, Wang, Pontil, Gong, Huang, and
  Tian}]{Peng_2016_CVPR}
Peng P, Xiang T, Wang Y, Pontil M, Gong S, Huang T, Tian Y (2016) Unsupervised
  cross-dataset transfer learning for person re-identification. In: The IEEE
  Conference on Computer Vision and Pattern Recognition (CVPR)

\bibitem[{Peng et~al.(2020)Peng, Xing, and Cao}]{penghybrid}
Peng P, Xing J, Cao L (2020) Hybrid learning for multi-agent cooperation with
  sub-optimal demonstrations. In: {IJCAI}, pp 3037--3043

\bibitem[{Perrone and Cooper(1992)}]{perrone1992networks}
Perrone MP, Cooper LN (1992) When networks disagree: Ensemble methods for
  hybrid neural networks. Tech. rep., BROWN UNIV PROVIDENCE RI INST FOR BRAIN
  AND NEURAL SYSTEMS

\bibitem[{Qi et~al.(2019)Qi, Wang, Huo, Zhou, Shi, and Gao}]{qi2019novel}
Qi L, Wang L, Huo J, Zhou L, Shi Y, Gao Y (2019) A novel unsupervised
  camera-aware domain adaptation framework for person re-identification. In:
  Proceedings of the IEEE International Conference on Computer Vision (ICCV),
  pp 8080--8089

\bibitem[{Ristani et~al.(2016)Ristani, Solera, Zou, Cucchiara, and
  Tomasi}]{DBLP:conf/eccv/RistaniSZCT16}
Ristani E, Solera F, Zou RS, Cucchiara R, Tomasi C (2016) Performance measures
  and a data set for multi-target, multi-camera tracking. In: IEEE {ECCV}
  Workshops

\bibitem[{Rousseeuw(1987)}]{rousseeuw1987silhouettes}
Rousseeuw PJ (1987) Silhouettes: a graphical aid to the interpretation and
  validation of cluster analysis. Journal of computational and applied
  mathematics 20:53--65

\bibitem[{Shen et~al.(2019)Shen, He, and Xue}]{shen2019meal}
Shen Z, He Z, Xue X (2019) Meal: Multi-model ensemble via adversarial learning.
  In: Proceedings of the AAAI Conference on Artificial Intelligence, vol~33, pp
  4886--4893

\bibitem[{Singh et~al.(2016)Singh, Hoiem, and Forsyth}]{singh2016swapout}
Singh S, Hoiem D, Forsyth D (2016) Swapout: Learning an ensemble of deep
  architectures. In: Advances in neural information processing systems, pp
  28--36

\bibitem[{Song et~al.(2018)Song, Wang, Zhang, Du, Zhang, Huang, and
  Wang}]{DBLP:journals/corr/abs-1807-11334}
Song L, Wang C, Zhang L, Du B, Zhang Q, Huang C, Wang X (2018) Unsupervised
  domain adaptive re-identification: Theory and practice. CoRR abs/1807.11334

\bibitem[{Spearman(1961)}]{spearman1961proof}
Spearman C (1961) The proof and measurement of association between two things.

\bibitem[{Srivastava et~al.(2014)Srivastava, Hinton, Krizhevsky, Sutskever, and
  Salakhutdinov}]{srivastava2014dropout}
Srivastava N, Hinton G, Krizhevsky A, Sutskever I, Salakhutdinov R (2014)
  Dropout: a simple way to prevent neural networks from overfitting. The
  journal of machine learning research 15(1):1929--1958

\bibitem[{Szegedy et~al.(2016)Szegedy, Vanhoucke, Ioffe, Shlens, and
  Wojna}]{szegedy2016rethinking}
Szegedy C, Vanhoucke V, Ioffe S, Shlens J, Wojna Z (2016) Rethinking the
  inception architecture for computer vision. In: Proceedings of the IEEE
  conference on computer vision and pattern recognition (CVPR), pp 2818--2826

\bibitem[{Tarvainen and Valpola(2017)}]{tarvainen2017mean}
Tarvainen A, Valpola H (2017) Mean teachers are better role models:
  Weight-averaged consistency targets improve semi-supervised deep learning
  results. In: Advances in neural information processing systems, pp 1195--1204

\bibitem[{Vinyals et~al.(2019)Vinyals, Babuschkin, Czarnecki, Mathieu, Dudzik,
  Chung, Choi, Powell, Ewalds, Georgiev et~al.}]{vinyals2019grandmaster}
Vinyals O, Babuschkin I, Czarnecki WM, Mathieu M, Dudzik A, Chung J, Choi DH,
  Powell R, Ewalds T, Georgiev P, et~al. (2019) Grandmaster level in starcraft
  ii using multi-agent reinforcement learning. Nature 575(7782):350--354

\bibitem[{Wan et~al.(2013)Wan, Zeiler, Zhang, Le~Cun, and
  Fergus}]{wan2013regularization}
Wan L, Zeiler M, Zhang S, Le~Cun Y, Fergus R (2013) Regularization of neural
  networks using dropconnect. In: International conference on machine learning,
  pp 1058--1066

\bibitem[{Wang and Zhang(2020)}]{wang2020unsupervised}
Wang D, Zhang S (2020) Unsupervised person re-identification via multi-label
  classification. In: Proceedings of the IEEE/CVF Conference on Computer Vision
  and Pattern Recognition, pp 10981--10990

\bibitem[{Wang et~al.(2018)Wang, Zhu, Gong, and Li}]{Wang_2018_CVPR}
Wang J, Zhu X, Gong S, Li W (2018) Transferable joint attribute-identity deep
  learning for unsupervised person re-identification. In: IEEE CVPR

\bibitem[{Wang et~al.(2020{\natexlab{a}})Wang, Lai, Huang, Gong, and
  Hua}]{wang2020camera}
Wang M, Lai B, Huang J, Gong X, Hua XS (2020{\natexlab{a}}) Camera-aware
  proxies for unsupervised person re-identification. arXiv preprint
  arXiv:201210674

\bibitem[{Wang et~al.(2020{\natexlab{b}})Wang, Zhang, Zheng, Liu, Sun, Li, and
  Wang}]{wang2020cycas}
Wang Z, Zhang J, Zheng L, Liu Y, Sun Y, Li Y, Wang S (2020{\natexlab{b}})
  Cycas: Self-supervised cycle association for learning re-identifiable
  descriptions. In: Computer Vision--ECCV 2020: 16th European Conference,
  Glasgow, UK, August 23--28, 2020, Proceedings, Part XI 16, Springer, pp
  72--88

\bibitem[{Wei et~al.(2018)Wei, Zhang, Gao, and Tian}]{Wei_2018_CVPR}
Wei L, Zhang S, Gao W, Tian Q (2018) Person transfer gan to bridge domain gap
  for person re-identification. In: IEEE CVPR

\bibitem[{Wu et~al.(2019)Wu, Yang, Liu, Liao, Lei, and Li}]{wu2019unsupervised}
Wu J, Yang Y, Liu H, Liao S, Lei Z, Li SZ (2019) Unsupervised graph association
  for person re-identification. In: Proceedings of the IEEE/CVF International
  Conference on Computer Vision, pp 8321--8330

\bibitem[{Xuan and Zhang(2021)}]{xuan2021intra}
Xuan S, Zhang S (2021) Intra-inter camera similarity for unsupervised person
  re-identification. In: Proceedings of the IEEE/CVF Conference on Computer
  Vision and Pattern Recognition, pp 11926--11935

\bibitem[{Yang et~al.(2020)Yang, Yan, Lu, Jia, Xie, Yu, Guo, Huang, and
  Gao}]{yang2020part}
Yang F, Yan K, Lu S, Jia H, Xie D, Yu Z, Guo X, Huang F, Gao W (2020)
  Part-aware progressive unsupervised domain adaptation for person
  re-identification. IEEE Transactions on Multimedia

\bibitem[{Yang et~al.(2021)Yang, Zhong, Luo, Cai, Lin, Li, and
  Sebe}]{yang2021joint}
Yang F, Zhong Z, Luo Z, Cai Y, Lin Y, Li S, Sebe N (2021) Joint noise-tolerant
  learning and meta camera shift adaptation for unsupervised person
  re-identification. In: Proceedings of the IEEE/CVF Conference on Computer
  Vision and Pattern Recognition, pp 4855--4864

\bibitem[{Ye et~al.(2017)Ye, Ma, Zheng, Li, and Yuen}]{ye2017dynamic}
Ye M, Ma AJ, Zheng L, Li J, Yuen PC (2017) Dynamic label graph matching for
  unsupervised video re-identification. In: Proceedings of the IEEE
  international conference on computer vision, pp 5142--5150

\bibitem[{Yuan et~al.(2019)Yuan, He, Zhu, and Li}]{yuan2019adversarial}
Yuan X, He P, Zhu Q, Li X (2019) Adversarial examples: Attacks and defenses for
  deep learning. IEEE transactions on neural networks and learning systems
  30(9):2805--2824

\bibitem[{Zeng et~al.(2020)Zeng, Ning, Wang, and Guo}]{zeng2020hierarchical}
Zeng K, Ning M, Wang Y, Guo Y (2020) Hierarchical clustering with hard-batch
  triplet loss for person re-identification. In: Proceedings of the IEEE/CVF
  Conference on Computer Vision and Pattern Recognition, pp 13657--13665

\bibitem[{Zhai et~al.(2020{\natexlab{a}})Zhai, Lu, Ye, Shan, Chen, Ji, and
  Tian}]{zhai2020ad}
Zhai Y, Lu S, Ye Q, Shan X, Chen J, Ji R, Tian Y (2020{\natexlab{a}})
  Ad-cluster: Augmented discriminative clustering for domain adaptive person
  re-identification. In: Proceedings of the IEEE/CVF Conference on Computer
  Vision and Pattern Recognition, pp 9021--9030

\bibitem[{Zhai et~al.(2020{\natexlab{b}})Zhai, Lu, Ye, Shan, Chen, Ji, and
  Tian}]{Zhai_2020_CVPR}
Zhai Y, Lu S, Ye Q, Shan X, Chen J, Ji R, Tian Y (2020{\natexlab{b}})
  Ad-cluster: Augmented discriminative clustering for domain adaptive person
  re-identification. In: IEEE/CVF Conference on Computer Vision and Pattern
  Recognition (CVPR)

\bibitem[{Zhai et~al.(2020{\natexlab{c}})Zhai, Ye, Lu, Jia, Ji, and
  Tian}]{zhai2020multiple}
Zhai Y, Ye Q, Lu S, Jia M, Ji R, Tian Y (2020{\natexlab{c}}) Multiple expert
  brainstorming for domain adaptive person re-identification. arXiv preprint
  arXiv:200701546

\bibitem[{Zhang et~al.(2019)Zhang, Cao, Shen, and You}]{zhang2019self}
Zhang X, Cao J, Shen C, You M (2019) Self-training with progressive
  augmentation for unsupervised cross-domain person re-identification. In:
  Proceedings of the IEEE International Conference on Computer Vision (ICCV),
  pp 8222--8231

\bibitem[{Zhang et~al.(2018)Zhang, Xiang, Hospedales, and Lu}]{zhang2018deep}
Zhang Y, Xiang T, Hospedales TM, Lu H (2018) Deep mutual learning. In:
  Proceedings of the IEEE Conference on Computer Vision and Pattern Recognition
  (CVPR), pp 4320--4328

\bibitem[{Zhao et~al.(2020)Zhao, Liao, Xie, Zhao, Zhang, and
  Shao}]{zhao2020unsupervised}
Zhao F, Liao S, Xie GS, Zhao J, Zhang K, Shao L (2020) Unsupervised domain
  adaptation with noise resistible mutual-training for person
  re-identification. In: European Conference on Computer Vision (ECCV),
  Glasgow, UK, pp 1--18

\bibitem[{Zheng et~al.(2021{\natexlab{a}})Zheng, Liu, He, Mei, Luo, and
  Zha}]{zheng2021group}
Zheng K, Liu W, He L, Mei T, Luo J, Zha ZJ (2021{\natexlab{a}}) Group-aware
  label transfer for domain adaptive person re-identification. In: Proceedings
  of the IEEE/CVF Conference on Computer Vision and Pattern Recognition, pp
  5310--5319

\bibitem[{Zheng et~al.(2015)Zheng, Shen, Tian, Wang, Wang, and
  Tian}]{Zheng_2015_ICCV}
Zheng L, Shen L, Tian L, Wang S, Wang J, Tian Q (2015) Scalable person
  re-identification: A benchmark. In: The IEEE International Conference on
  Computer Vision (ICCV)

\bibitem[{Zheng et~al.(2016)Zheng, Yang, and Hauptmann}]{zheng2016person}
Zheng L, Yang Y, Hauptmann AG (2016) Person re-identification: Past, present
  and future. arXiv preprint arXiv:161002984

\bibitem[{Zheng et~al.(2021{\natexlab{b}})Zheng, Tang, Teng, Ge, Liu, Qin, Qi,
  and Chen}]{zheng2021online}
Zheng Y, Tang S, Teng G, Ge Y, Liu K, Qin J, Qi D, Chen D (2021{\natexlab{b}})
  Online pseudo label generation by hierarchical cluster dynamics for adaptive
  person re-identification. In: Proceedings of the IEEE/CVF International
  Conference on Computer Vision, pp 8371--8381

\bibitem[{Zheng et~al.(2017)Zheng, Zheng, and Yang}]{Zheng_2017_ICCV}
Zheng Z, Zheng L, Yang Y (2017) Unlabeled samples generated by gan improve the
  person re-identification baseline in vitro. In: IEEE ICCV

\bibitem[{Zhong et~al.(2017)Zhong, Zheng, Cao, and Li}]{Zhong_2017_CVPR}
Zhong Z, Zheng L, Cao D, Li S (2017) Re-ranking person re-identification with
  k-reciprocal encoding. In: IEEE CVPR

\bibitem[{Zhong et~al.(2018)Zhong, Zheng, Li, and
  Yang}]{DBLP:conf/eccv/ZhongZLY18}
Zhong Z, Zheng L, Li S, Yang Y (2018) Generalizing a person retrieval model
  hetero- and homogeneously. In: ECCV, pp 176--192

\bibitem[{Zhong et~al.(2019{\natexlab{a}})Zhong, Zheng, Luo, Li, and
  Yang}]{Zhong_2019_CVPR}
Zhong Z, Zheng L, Luo Z, Li S, Yang Y (2019{\natexlab{a}}) Invariance matters:
  Exemplar memory for domain adaptive person re-identification. In: IEEE CVPR

\bibitem[{Zhong et~al.(2019{\natexlab{b}})Zhong, Zheng, Zheng, Li, and
  Yang}]{DBLP:journals/tip/ZhongZZLY19}
Zhong Z, Zheng L, Zheng Z, Li S, Yang Y (2019{\natexlab{b}}) Camstyle: {A}
  novel data augmentation method for person re-identification. IEEE TIP
  28(3):1176--1190

\bibitem[{Zhong et~al.(2020)Zhong, Zheng, Luo, Li, and
  Yang}]{zhong2020learning}
Zhong Z, Zheng L, Luo Z, Li S, Yang Y (2020) Learning to adapt invariance in
  memory for person re-identification. IEEE Transactions on Pattern Analysis
  and Machine Intelligence

\bibitem[{Zhou et~al.(2019)Zhou, Yang, Cavallaro, and Xiang}]{zhou2019omni}
Zhou K, Yang Y, Cavallaro A, Xiang T (2019) Omni-scale feature learning for
  person re-identification. In: Proceedings of the IEEE/CVF International
  Conference on Computer Vision, pp 3702--3712

\bibitem[{Zou et~al.(2020)Zou, Yang, Yu, Kumar, and Kautz}]{zou2020joint}
Zou Y, Yang X, Yu Z, Kumar BV, Kautz J (2020) Joint disentangling and
  adaptation for cross-domain person re-identification. In: Computer
  Vision--ECCV 2020: 16th European Conference, Glasgow, UK, August 23--28,
  2020, Proceedings, Part II 16, Springer, pp 87--104

\end{thebibliography}

\end{document}